\pdfoutput=1
\documentclass[11pt]{article}

\usepackage[final]{acl}

\usepackage{times}
\usepackage{latexsym}
\usepackage{cleveref}
\usepackage{url}
\usepackage{paralist}
\usepackage{enumitem}

\usepackage[T1]{fontenc}
\usepackage[utf8]{inputenc}

\usepackage{microtype}
\usepackage{xspace}
\usepackage{inconsolata}
\usepackage{booktabs}

\newcommand{\conflict}{\texttt{conflict}\xspace}

\newcommand{\hero}{\texttt{hero}\xspace}
\newcommand{\villain}{\texttt{villain}\xspace}
\newcommand{\victim}{\texttt{victim}\xspace}
\newcommand{\focus}{\texttt{focus}\xspace}
\newcommand{\story}{\texttt{cultural story}\xspace}

\newenvironment{myquote}%
  {\list{}{\leftmargin=0.1in\rightmargin=0.1in}\item[]}%
  {\endlist}

\usepackage{graphicx}

\title{Narrative Media Framing in Political Discourse}

\author{Yulia Otmakhova \quad Lea Frermann\\
  School of Computing and Information Systems, \\The University of Melbourne \\\\
  \texttt{\{y.otmakhova, l.frermann\}@unimelb.edu.au}}
\begin{document}
\maketitle
\begin{abstract}

Narrative frames are a powerful way of conceptualizing and communicating complex, controversial ideas, however automated frame analysis to date has mostly overlooked this framing device. In this paper, we connect elements of narrativity with fundamental aspects of framing, and present a framework which formalizes and operationalizes such aspects. We annotate and release a data set of news articles in the climate change domain, analyze the dominance of narrative frame components across political leanings, and test LLMs in their ability to predict narrative frames and their components. Finally, we apply our framework in an unsupervised way to elicit components of narrative framing in a second domain, the COVID-19 crisis, where our predictions are congruent with prior theoretical work showing the generalizability of our approach.\footnote{We release our code, data and annotations at \url{https://github.com/julia-nixie/narratives}.}

\end{abstract}

\section{Introduction}
\label{sec:introduction}

Narrative framing is a type of media framing that uses elements of narrativity to highlight some aspects of a complex issue and condense it into a simplified ``story'' that promotes a particular interpretation~\cite{crow2016media}. These elements of storytelling, such as representing an issue through the lens of stakeholders and conflicts rather than direct description of the facts, make narrative frames a highly effective device, particularly in the context of contested issues such as climate change~\citep{daniels2009narratives,rodrigo2019talking}.

\begin{figure}[t]
    \centering
    \includegraphics[width=1\linewidth]{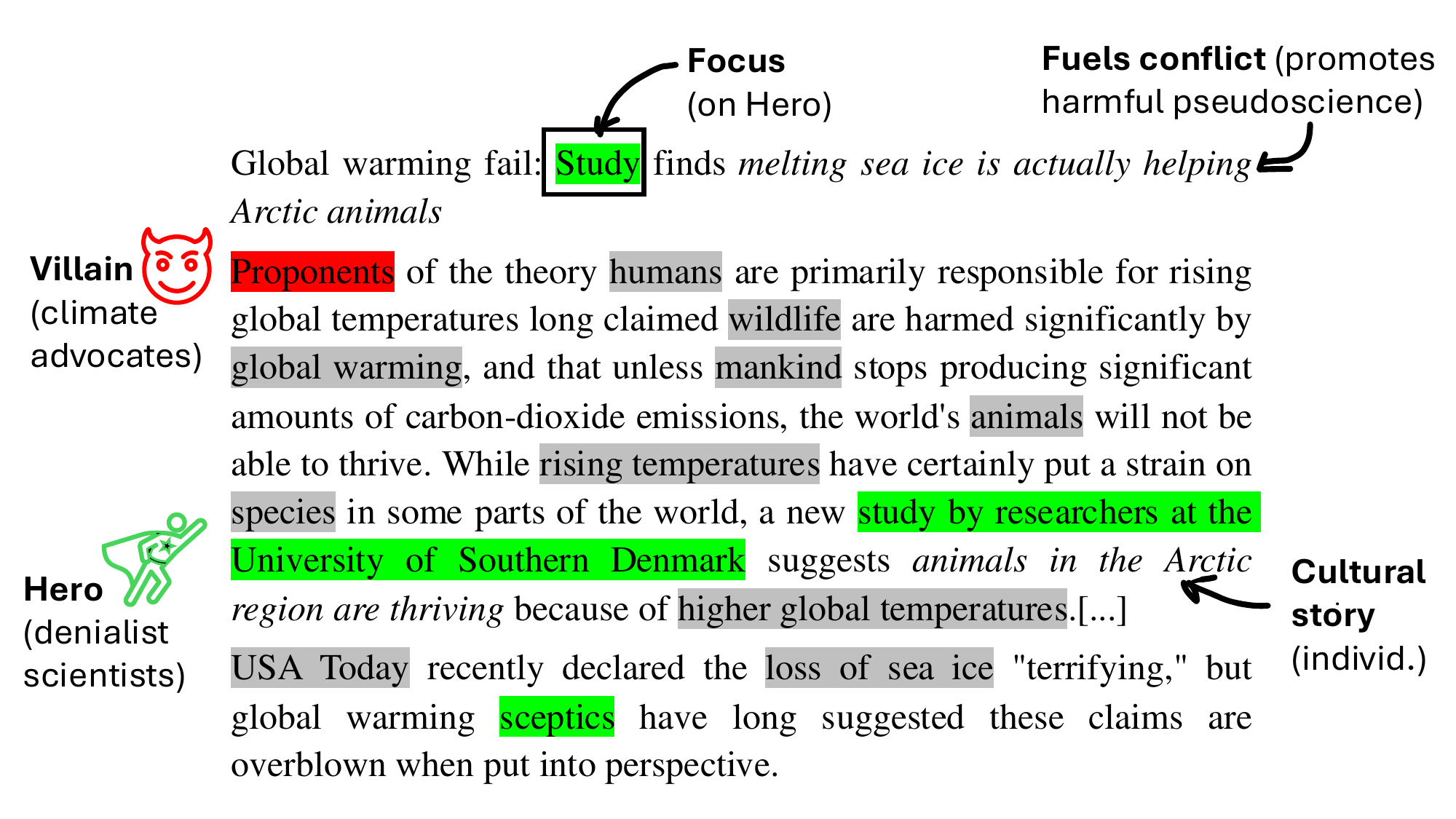}
    \caption{An except from a news article, with \hero marked in green and \villain in red. Entities that are not main characters are grayed out. The box shows the focal character (here, \hero). The phrases in italic are cues which show that the article has an \textit{individualistic} cultural story (``the nature can fix itself'') and that it \textit{fuels conflict} by actively promoting bad science.}
    \label{fig:hvv}
\end{figure}

Narrative framing can draw the reader's attention to specific, nuanced aspects of an issue and instill a very precise interpretation that differs from the ``default'' reading inferrable from its generic or issue-specific frame. To give an example, the text in \Cref{fig:hvv} frames the topic of climate change through a ``Polar Bear'' issue-specific frame~\cite{bushell2017strategic}
% \footnote{None of the widely adopted generic frames such as by \citet{boydstun2013identifying} or~\citet{semetko2000framing} is readily applicable to this text, which shows their interpetative limitations.} 
which describes the negative effects of climate change on animals (\textit{rising temperatures have certainly put a strain on species}). However, this is not the actual message of the text: it depicts climate scientists as incorrect, while presenting pseudo-scientists from a hero-like angle. It uses devices of narrative framing to replace the default interpretation arising from a topic-like frame (``animals are victims of climate change and humans are villains'') with an opposing idea that animals are doing fine and scientists who claim otherwise mislead the general public.

While the importance of framing narratives for the communication and perception of news has been widely recognized in the social sciences~\cite{shanahan2011narrative}, automatic framing analysis still mostly conceptualizes frames in a topic-like fashion~\citep{ali2022survey,otmakhova-etal-2024-media}. Recent work in NLP has studied elements of narratives such as characters or events in news reporting~\citep{stammbach-etal-2022-heroes,frermann-etal-2023-conflicts,gehring2023analyzing,zhao-etal-2024-media,das-etal-2024-media}. However, these studies are either topic-specific, or lose the link to the core mechanism of framing: For framing to occur, an {\bf ambivalent} issue must be present \cite{sniderman2004structure} and described in a way that \textbf{evokes a larger interpretative context} (schema) which goes beyond information directly inferrable from the text \citep{scheufele2010spreading}. As such, beyond the issue itself (aka \citet{entman1993framing}'s ``problem statement''), there are other important mechanisms that turn a message into a frame, such as alluding to the \textbf{conflict} that led to the problem (=its cause), making a \textbf{moral evaluation}, or suggesting a \textbf{resolution} to it. We present a general formalization of narrative framing that comprises all these aspects.

We do so by integrating elements of narratology with social and media studies on narrative framing, to establish a framework which allows to identify and structurally represent narrative frames and to distinguish superficially similar frames from one another. First, our framework distinguishes character roles, and we show how issue \textbf{ambivalence} arises when actors in an article are assigned an archetypical role (Hero, Villain, Victim) and one of the roles is drawn into focus. This emphasis in turn evokes a {\bf moral evaluation} of the personas in the article.
% \footnote{In  \Cref{fig:hvv} climate activists receive the role of a Villain, while denialist scientists are Heroes. The narrative focuses on the heroes' ``contributions".}
% Such assignment of morally loaded archetype roles also prescribes a particular \textbf{evaluation} of personas in the article. 
As a second component, we position each character in terms of exacerbating or resolving the core {\bf conflict} of the article.
% broad relationships between characters to a general definition of \textbf{conflict} and \textbf{resolution} 
% (the article in \Cref{fig:hvv} exacerbates the cause of climate crisis rather than fuels its resolution). 
Finally, to link the presentation in the article to the \textbf{wider set of associations and beliefs} already existing in the receiver's perception \citep{nelson1997toward}, we link our narrative frames to established ``cultural stories'' that define the attitude towards external control and the sense of unity with the group \citep{thompson2018cultural}. Figure~\ref{fig:hvv} illustrates the three components with an example.
% The text above is an example of individualistic cultural story, which implies that no control is necessary and the society does not need to act as one.

We apply our framework to analyze media framing of two distinct public issues -- climate change and COVID-19. In particular, we make the following contributions:

\begin{enumerate}
    \item We define elements of narrativity that are essential for narrative framing and are aligned with the definition of a media frame and show that our framework applies across topics (climate change and COVID-19) and domains (news articles and political speeches).
    %, and that it can capture new, emergent narratives and differentiate them from previously reported in the literature.
    \item We show that framework enables reliable and effective annotation of narrative frames in the news, and improves the automatic detection of narratives.
    \item We release a corpus of articles about climate change annotated with narrative frames, and use it to analyze the distribution of different narrative frames across political leanings.
    \item We test a range of LLMs on their ability to automatically predict the components of our framework, leaving room for improvements.
\end{enumerate}

\section{Background}

\paragraph{Narratives in political communication}
Following \citet{fisher1984narration}'s seminal paper coining the term ``homo narrans'' to illustrate the importance of storytelling for society, narratives in political communication have attracted substantial research attention (see also~\citet{bennett1985toward,patterson1998narrative}), exposing its effects from a critical vehicle in deliberative democracy~\cite{boswell2013and} to its use persuasive device~\cite{skrynnikova2017power}. Similar to the concept of framing in general, a principled and empirically testable definition of ``narrative framing'' has long been lacking. However, recent work has progressed in developing frameworks that are testable and amenable to computational modeling~\cite{shenhav2005thin,robert2014fundamental}, most prominently the Narrative Policy Framework (NPF;~\citet{jones2023narrative}), which we build on in this work. The NPF defines a set of generalizable structural elements in political/policy narratives, including characters, settings, plot and moral evaluation which it uses to characterise the operation of narratives on the individual, group and cultural level. The NPF has been particularly instrumental in studying  climate change narratives~\cite{flottum2017narratives}, and identifying dominant narratives in the discourse~\cite{bushell2017strategic,bevan2020climate}. This paper adapts the elements of the NPF into a structured framework designed to support automated prediction.

\paragraph{Narrative framing and NLP}
Narrative framing intersects the concepts of storytelling and {framing}, i.e., the presentation of information in a way to evoke a specific association in the audience. Automatic narrative understanding has attracted substantial attention in NLP~\citep{piper-etal-2021-narrative}; however, it has focused mostly on fictional narratives~\citep{bamman-etal-2014-bayesian,iyyer-etal-2016-feuding}, personal narratives in social media~\cite{lukin-etal-2016-personabank,shen-etal-2023-modeling}, or specific elements such as event chains~\cite{chambers-jurafsky-2009-unsupervised}. Few works have considered the intersection of stories and framing~\cite{levi-etal-2022-detecting}, or used elements of narrativity such as events to improve on topic-focussed framing analysis \citep{das-etal-2024-media,zhao-etal-2024-media}.

While narrative framing research in the social sciences is strongly grounded in the NPF, this framework is yet to gain recognition in NLP approaches. Closest to our work are \citet{stammbach-etal-2022-heroes} and \citet{frermann-etal-2023-conflicts} who study some narrative elements of framing devices (such as entities framed as heroes or victims), but do not model full narrative frames. In addition to entities \citet{gehring2023analyzing} model relationships between them such as ``harm'' or ``protect''; however, their approach does not map the identified elements back to more high-level (narrative) frames. %Overall, our approach is the first to go beyond storytelling elements to policy narrative structures used as framing devices.

\section{Components of narrative framing}
\label{sec:components}

We motivate our three core components which define a narrative frame. Each component contributes to the framing mechanism, by resolving the ambivalence through assigning moral evaluation to stakeholders (Characters), capturing the conflict and resolution aspect of a frame (Conflict and resolution), and evoking a wider set of cognitive schemata and cultural associations (Cultural stories).

\subsection{Characters}
\label{sec:character_annotation}

Characters and their prototypical roles have been studied extensively in narratology (starting from formalist and structuralist approaches such as \citet{propp1968morphology} and \citet{greimas1987meaning}), and were adopted as a simplified \hero, \villain, and \victim (HVV) triad by social sciences as part of Narrative Policy Framework (NPF) \citep{shanahan2018conduct}\footnote{In NLP, character (or agent) identification has attracted substantial attention, both from the narratology side (see \citep{piper-etal-2021-narrative}) and, less extensively, from the NPF angle \citep{frermann-etal-2023-conflicts}.}. In particular, the NPF prescribes that a narrative frame should include {\bf at least one prototypical character}, i.e. one or more HVV roles should be filled by a prominent entity. 
By assigning an entity to a particular role, we resolve the issue ambivalence by conveying our moral judgment of that entity, as required by Entman's definition of a frame \citep{entman1993framing}. Essentially, the reader's interpretation of the article depends on whether a particular entity (say, \textit{climate advocates} as in \Cref{fig:hvv}) is framed as a \hero (their actions are evaluated as beneficial), a \villain (as in \Cref{fig:hvv}), or \victim (of criticism or attacks by denialists).

Often there are multiple candidate entities for each HVV role in a text. We follow narratology approaches in distinguishing between main characters and other entities \citep{jahan-finlayson-2019-character}, and use the {\it single most central character} fulfilling the respective role to represent a narrative frame. \Cref{fig:hvv} illustrates this, where the main characters are highlighted in color, while less central entities are grayed out. Moreover, to be able to compare instances of a particular narrative frame across texts with different people and events, we abstract away from specific characters to the stakeholder categories (common people, elites, etc.) they represent. The taxonomy of such stakeholders can either be inherited from the literature (as we do in \Cref{sec:supervised}) or derived from a corpus in a data-driven way (as demonstrated in \Cref{sec:unsupervised}).

To fully differentiate narratives, in addition to assigning characters to roles, it is necessary to identify the \textbf{focus} on either \hero, \villain, or \victim, which results in ``heroic'', ``blaming'', and ``victimizing'' narrative frames, respectively. For example, two distinct narrative frames can both frame \textit{climate activists} as a \hero and \textit{government} as a \villain, but focus either on criticizing the government (``blaming'') or praising the efforts of activists in opposing it (``heroic'') -- resulting in very different messaging.\footnote{Examples from \citet{bevan2020climate}, see narrative frames ``The collapse is imminent'' vs ``You're destroying our future'' in \Cref{app:narratives}.}

\subsection{Conflict and resolution}
\label{sec:resolution_annotation}

Conflict/resolution\footnote{Here we define {it conflict} as an underlying cause of an issue which characters strive to either escalate or resolve, rather than a driving force of a plot~\citep{prince2003dictionary} or breaking point in its canonicity~\citep{bruner1991narrative}, as understood in narratology.} is a central element of a narrative frame.  It encapsulates the ``plot'' element of the NPF, and links into Entman's \citep{entman1993framing} criteria of framing which state that, among its other functions, a frame can point to the cause of the issue and its underlying conflict, or prescribe a resolution.
Accordingly, we conceptualize conflict and resolution as a four-way distinction: the characters assigned \hero and \victim roles in a narrative frame can either \textit{fuel conflict} (perform actions that cause or exacerbate the issue), \textit{fuel resolution} (perform actions that help to resolve the issue), \textit{prevent conflict} (oppose actions that cause or exacerbate the issue), or \textit{prevent resolution} (oppose actions that help to resolve the issue). 

In NLP, relations between characters have a long history of research \citep{agarwal2010automatic,shahsavari2020conspiracy}, including studies which specifically looks at conflicts \citep{han-etal-2019-permanent,olsson-etal-2020-text}. In comparison, our framework abstracts away from specific (and often sparse) entity relations and combines the attitude towards the issue (pro-conflict vs pro-resolution) with the level of intentionality and direct expression of that attitude (i.e.~actively perform actions that support one's side, or oppose the actions of the other side). This definition of conflict/resolution based on abstract categories rather than on specific actions or events renders our approach generalizable across topics, as we show in \Cref{sec:supervised,sec:unsupervised}.

\subsection{Cultural stories}
\label{sec:culture_stories}

\begin{figure}
    \centering
    \includegraphics[width=0.7\linewidth]{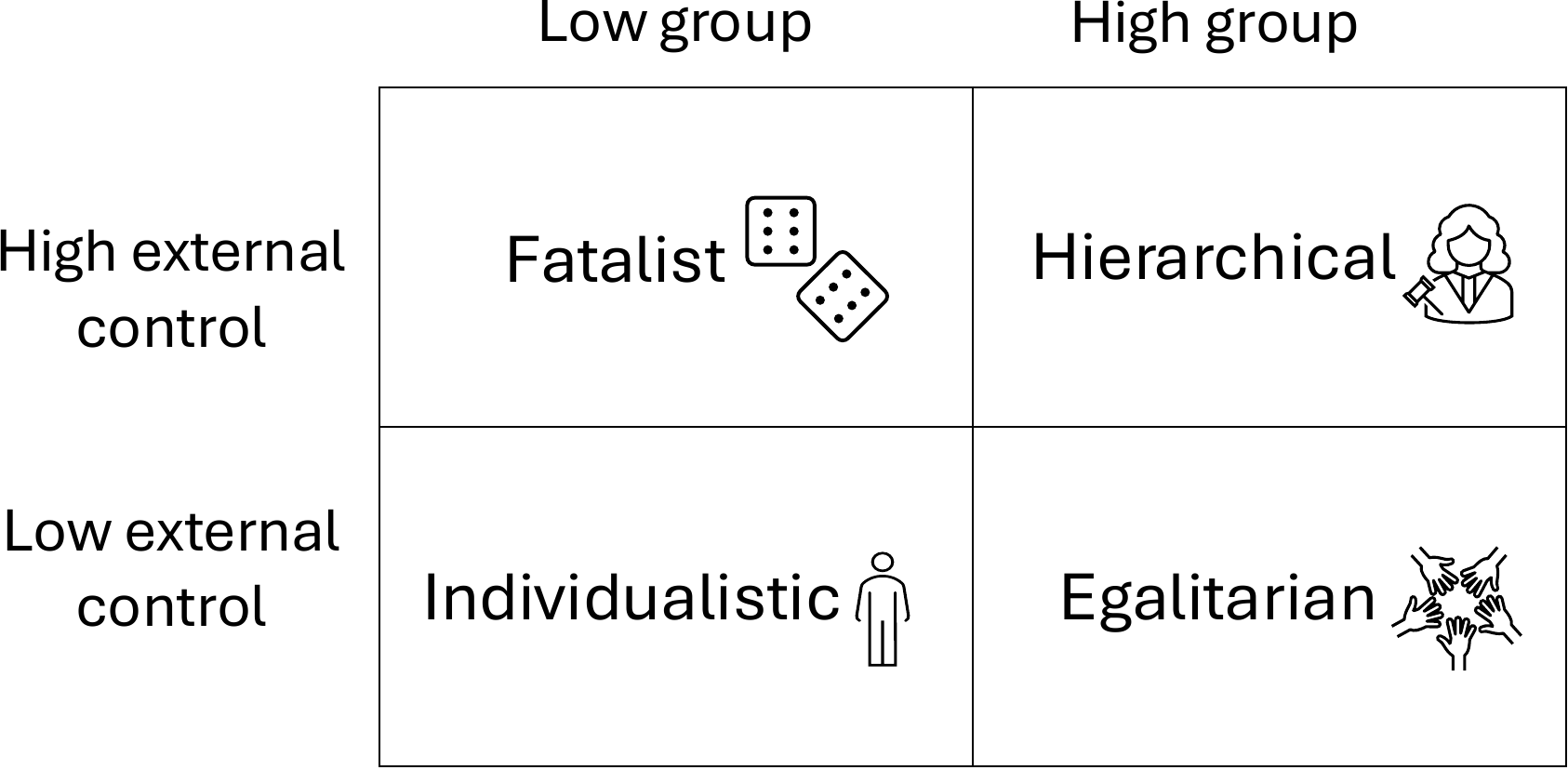}
    \caption{Cultural stories across dimensions of \textbf{external control} (grid) and belonging to a \textbf{group}}
    \label{fig:stories}
\end{figure}

Frames are distinguished from ``unframed'' types of communication by their ability to 
evoke a wider set of concepts, associations and judgments which already exist in the audience's perception~\citep{scheufele2010spreading}. Narrative frames do this by mapping a particular combination of characters and conflict/resolution to one of four larger schemata of interpretation, which in social studies are referred to as {\bf cultural value stories} \citep{thompson2018cultural}.\footnote{Thus, all narrative frames are stories, i.e.~contain elements of narrativity such as characters and plot (reduced to conflict and resolution). However, not all stories can be used as narrative frames: in order to so, they need to map to a broader, pre-existing context dictated by a cultural story.} Cultural stories define to what degree our actions are controlled by external factors and by the sense of belonging to a group \citep{douglas2007history}. Depending on the combination of these two factors, a narrative frame can be \textit{fatalist} (where people are at the mercy of forces outside their control, such as natural disasters or fate), \textit{hierarchical} (where people are bound by social prescriptions and external control, such as government), \textit{individualistic} (where social ties are loose and people reject the necessity of external control), or \textit{egalitarian} (where people take collective action, opposing external control) (\Cref{fig:stories}). Cultural stories have been shown to directly affect public behavior: as an example, individualist and egalitarian stories have been linked to worse survival rates than a hierarchical story during the COVID-19 onset \citep{guss2021individualism}.

To the best of our knowledge, cultural stories, or more generally cognitive and cultural schemata aiding interpretation, have not been explored in NLP. However, many NLP studies \citep{finlayson2012learning,tangherlini2020automated} draw upon related concepts of narrative archetypes as overarching, culturally repetitive plots or narrative elements~\citep{northrop1957anatomy,propp1968morphology}. In contrast, we focus on framing and its link to a well-defined space of cultural values which have been shown to affect perception and behavior.

\section{Narrative Framing of Climate Change}
\label{sec:supervised}

In the remainder of this paper we apply our framework to perform narrative framing analysis on two topics: climate change and COVID-19. First, in this section, we use it to manually annotate the three components of narrative frames in news articles, and then map them to established narrative frames in the climate change domain. We use this corpus to evaluate the ability of multiple LLMs to predict narrative frame components and the frames themselves. Then, in \Cref{sec:unsupervised} we show how the framework can be generalized to a domain without an established repertoire of narrative frames (politician's speeches on COVID-19), where the goal is to \textit{discover} frames rather than classify them. 
% Specifically, we present an exploratory analysis of politician's speeches during the onset of COVID-19 and show how our framework surfaces insights consistent with prior analyses.

% \subsection{Supervised approach: climate domain}

\subsection{Data selection and annotation}

\textbf{Article selection.} We manually annotate 100 articles randomly selected from an existing dataset of news stories on the topic of climate change~\citep{frermann-etal-2023-conflicts}, originating from a range of US media outlets from different political leanings published between 2017 and 2019.\footnote{Due to space constraints, this paper covers only our US centric analysis. However, we also release an annotated dataset of 100 Australian climate-focused articles from 2024, which surfaced new narratives with different combinations of elements to previously recorded ones. The dataset together with its analysis is available at \url{https://github.com/julia-nixie/narratives}.}
The articles are fairly evenly distributed across political leanings to ensure that the dataset contains a variety of narratives coming from different political groups. Detailed statistics are in \Cref{app:dataset_stats}.

\noindent\textbf{Annotation process}. We use the example in \Cref{fig:hvv} to explain the steps of the annotation process.
First, we identify candidate entities for the \hero, \villain, and \victim roles, and select at most one main character per role (as described in \Cref{app:deriving}). In our example, since \hero, \villain, and \victim should align with the article's perspective, we remove potential victims like \textit{animals} since the author believes they actually benefit from higher temperatures. Then, using an established taxonomy of stakeholder categories for the climate change domain (\citet{frermann-etal-2023-conflicts}; details in \Cref{app:stakeholder}), we map the text spans that represent characters to labels indicating general classes of actors. Thus, we arrive at \textit{science experts} (from the skeptics side) as \hero and \textit{environmental activists} as \villain. To determine the focus, we rely on discourse structure of newspaper articles, namely the inverted pyramid where the most important content is presented first, and the relative proportion of text devoted to the different roles. Since the title highlights the research of climate skeptics and much of the article's content is devoted to describing it, we determine that the \focus is on the \hero. Next, since the article explicitly promotes dubious science harmful to the climate (rather than only criticizing actions of climate activists), it fuels \conflict. Finally, as the article implies that nature is resilient and no actions are necessary, it corresponds to an \textit{individualistic} cultural story.  \Cref{app:annotation} provides more details on the annotation process, instructions, and quality assurance.

We apply this process to annotate the structure of the narrative frames in our corpus of 100 climate change articles, and use the same framework to determine the components of known narrative frames from the climate change literature~\cite{bushell2017strategic,bevan2020climate,lamb2020discourses}. Then, we map the article structures to the structures of known narrative frames to arrive at the final narrative frame label for the article. For example, the structure of the article in~\Cref{fig:hvv} points to a denialist narrative frame ``No need to act". Overall, this element-wise mapping between the articles and the theoretical literature resulted in defining 16 structurally distinct narrative frames, which are described in detail in~\Cref{app:narratives}. 

\noindent\textbf{Annotation quality.} Based on the component-wise annotation process described above we achieve reliable (in terms of Krippendorf $\alpha$ among all annotators) and very strong (in terms of agreement with an expert) inter-annotator agreement on all elements of the framework.  In particular, we achieve a Krippendorf $\alpha$ agreement of 0.76 for \hero, 0.67 for \villain, and 0.81 for \victim between four annotators, and  Krippendorf $\alpha$ of 0.78 for \focus, 0.82 for \conflict, and 0.80 for Cultural Story between two annotators. 

Since each narrative frame is derived from a unique combination of its elements, the reliable annotation of narrative frame components also ensures a more reliable annotation of resulting narratives than choosing them based on their description only. To demonstrate that, we compare the agreement between narrative labels derived from their components against agreement in a setting where annotators chose one out of 16 narratives directly based on their descriptions~(\Cref{app:str_annotation}). The former leads to substantially higher agreement (63\% vs 37\%). 
% Thus, annotating the components one-by-one, rather than choosing the narrative from a list of descriptions, reduces complexity of the task and helps the annotators make more reliable choices. 
Furthermore, the component-wise annotation was significantly faster than direct labeling due to the reduced cognitive load of annotators. 
% Thus, we believe this study highlights a promising approach to reliable annotation of complex, multi-class tasks where the classes are difficult to differentiate based on their descriptions only.

\noindent{\bf Final dataset.} The final dataset contains 16 climate change narrative frames, as well as their components, and covers the majority of narrative frames mentioned in social studies literature. It includes frames that are similar on the surface, but differ in structure and thus can be used as a challenging test set for narrative frame detection in this domain. Full dataset statistics regarding the distribution of narrative frames and their components are provided in \Cref{app:data_stats}. 

 \subsection{Dataset Analysis}
 \label{ssec:analysis}
We analyze how the annotated narrative frames and their components vary across articles from different political leanings, and their alignment with more commonly used generic frames.

\begin{figure}[t]
    \centering
    \includegraphics[width=1\linewidth]{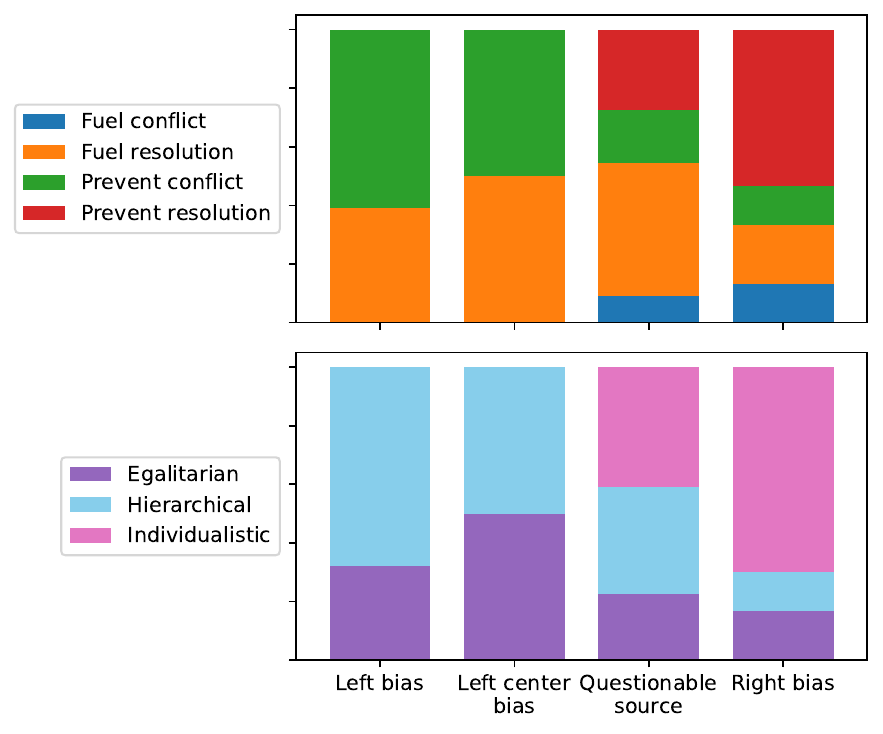}
    \caption{Distribution of \conflict and \story values across political leanings}
    \label{fig:story_conflict}
\end{figure}

\noindent\textbf{Narrative frames across political leanings.} 
Individual narrative components strongly associate with specific political leanings of news outlets: The overwhelming majority of right-bias articles are framed as {\it preventing resolution} (of the climate crisis) and exemplify the {\it individualistic} cultural story (\Cref{fig:story_conflict}), while these values do not appear at all for left-bias and left-center outlets. Moreover, right-bias outlets frame scientists as heroes much more often than other sources (Appendix \Cref{fig:hero_leaning}): they often quote `'fake experts'' in support of their anti-climate statements (as in ~\Cref{fig:hvv}). The overall narrative labels are more distributed across political leanings (see \Cref{fig:narr_leaning}). Taken together, this shows that dissecting narrative frames into meaningful components affords more nuanced insights in relation to external metadata.

\begin{figure}[t]
    \centering
    \includegraphics[width=1\linewidth]{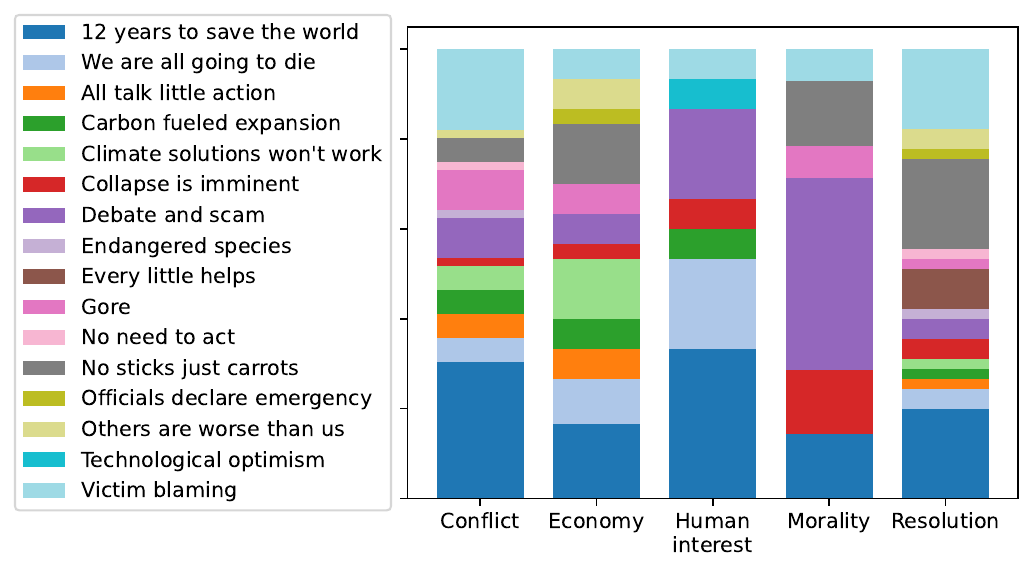}
    \caption{Narrative frames vs generic frames}
    \label{fig:generic_frames}
\end{figure}

\noindent\textbf{Narrative frames vs generic frames.} The 100 articles in our narrative framing corpus were previously annotated with five generic frames originally defined by \citet{semetko2000framing} (Conflict, Economic, Human Interest, Morality and Resolution). We intersect our labels with those generic frames to study the correlation of generic frames and narrative frames. 
% To support the hypothesis that the topical, token-focused frames such as generic and issue frames cannot capture the variety of intent expressed by narrative frames, we show the distribution of narrative frames across generic frames (as annotated by human annotators in \citep{frermann-etal-2023-conflicts} where we derive the articles from). 
\Cref{fig:generic_frames} shows little systematic correlation between generic and narrative frames: an article with a particular generic frame can have a variety of different narratives, and vice versa. For example, an "Economy"-framed article can emphasize the importance of fossil fuels ("Carbon fueled expansion" narrative), or focus on the economical effects of climate change ("12 years to save the world"). These observations align with theoretical works which showed that the same generic frame can have different intents (cf. \citet{bushell2017strategic,shanahan2007talking} for "Economy" frame), and highlights the added insights afforded by a narrative-focused frame analysis on top of generic emphasis frames.

\subsection{Automatic Prediction of Narrative Frames and Components}
\label{sec:experiments}

We use our dataset to test narrative frame prediction capabilities of LLMs. We define the following predictive tasks, given the full article text as input:
\begin{itemize}[leftmargin=0em, itemindent=1em]
\item  Choosing the stakeholder category for \hero, \villain, and \victim (separately for each character type) as one of 10 classes (\textit{government}, \textit{climate activists}, etc.; see \Cref{sec:character_annotation}). To choose a stakeholder correctly, a model needs to perform several steps: determining if an entity is framed as a \hero, \villain, or \victim, aggregating mentions of entities across the text to determine which of potential candidates is a main \hero, \villain, or \victim; and finally determining to which stakeholder category this character belongs.

\item Predicting the \focus entity out of 3 classes (\hero, \villain, or \victim). This task tests if a model can determine if the narrative frame is ``heroic'', ``blaming'', or victim-centered (see \Cref{sec:character_annotation}).

\item Predicting \conflict out of 4 classes (\textit{fuel conflict, fuel resolution, prevent conflict, prevent resolution}, see \Cref{sec:resolution_annotation}). A model needs to identify the general intent of the narrative frame (if it pushes towards resolution of the crisis, or exacerbates it), and the article's strategy to do so (by supporting one side or by criticizing the opposite side).

\item Predicting a \texttt{cultural story}\xspace out of 3 classes (\textit{individualistic, egalitarian, or hierarchical}, see~\Cref{sec:culture_stories})\footnote{Though \Cref{sec:culture_stories} introduces 4 cultural stories~\cite{jones2014cultural}, the fatalist story is not present in our data set so we exclude it from experiments for fair evaluation.}. To do so, the model needs to identify if the text implies collective vs individualistic action, and approval or disapproval of external control (such as from the government).

\item Directly predicting one of 16 \texttt{narrative frames}\xspace given an article, based on their short descriptions sourced from the social studies literature (full list in \Cref{app:narratives}).
\end{itemize}

\begin{table*}[ht]
\begin{small}
    \begin{center}
    \begin{tabular}{l|ccccccc}
         & Hero (10) & Villain (10) & Victim (10) & Focus (3) & Conflict (4) & Story (4) & Narrative (16) \\
         \toprule
         Baseline &  0.079	& 0.08	& 0.135	& 0.231	& 0.135	& 0.19	& 0.021 \\\midrule
         GPT4o &  0.325	& 0.454	& 0.266	& 0.656	& 0.332	& 0.574	& 0.258\\
         o1 &  0.363\textsuperscript{**}	& 0.527\textsuperscript{*}	& 0.455\textsuperscript{*}	& 0.718\textsuperscript{**}	& \textbf{0.549}	& \textbf{0.595}	& 0.330\textsuperscript{*}\\
         Mixtral &  0.237	& 0.073 & 0.257	& 0.402	& 0.353	& 0.431	& 0.171\\
         Llama &  0.271	& 0.156	& 0.336	& 0.568	& 0.379	& 0.449 & 0.181\\
         Gemini & 0.326	& 0.292	& 0.230	&  0.635	& 0.361 &	0.482  &	0.319 \\
         Sonnet & \textbf{0.353}	& \textbf{0.530}	& \textbf{0.469}	&  \textbf{0.688}	& 0.399 &	0.561  &	\textbf{0.339} \\
         \bottomrule
    \end{tabular}
    \end{center}
    \end{small}
    \caption{Zero-shot performance of 6 models across 7 narrative understanding tasks (macro-averaged F1). The number in brackets after the task's name indicates the number of classes in it. The baseline is calculated by using the most frequent label for the task as a predicted class. Results that had high (over 0.02) or very high (over 0.05) standard deviation across 5 runs are marked with * and ** respectively. The best performing models (considering variance) are in bold.} 
    \label{tab:zero_shot}

\end{table*}

\subsubsection{Models and prompts}

We use our tasks to test narrative frame prediction of 5 base LLMs of different size (GPT4o, Mixtral, Llama, Gemini and Claude Sonnet), and one reasoning LLM (o1).\footnote{Versions used: gpt-4o-2024-11-20, o1-preview-2024-09-12, Mixtral-8x7B-Instruct-v0.1, Llama-3.1-8B-Instruct, gemini-1.5-flash, Sonnet 3.5. Model sizes are provided in \Cref{app:size}.} We set temperature=0 (except for o1 which does not allow to control generation) to ensure deterministic outputs. We perform each experiment 5 times to ensure there is no substantial variance in the results. With the exception of o1, which shows high variance on most tasks, models have zero or near-zero variance across runs, which allows to compare averaged results.\footnote{When comparing with o1-preview-2024-09-12, we used the worst results rather than average to account for large variance.} 

The prompts used for each of the tasks are listed in \Cref{app:prompts}. The text of the prompts is based on descriptions of particular classes (stakeholders, culture stories, narrative frames etc.) in the social science literature. Prompts for HVV characters are domain-specific, i.e.~they are based on a list of entities important for the climate change domain (we show how to generalize this approach by creating such list automatically in \Cref{sec:unsupervised}). Conversely, prompts for Focus, Conflict, and Cultural story tasks are domain-agnostic and describe the classes in general terms (e.g., \textit{INDIVIDUALISTIC: this story assumes that the situation cannot be controlled externally, and no group actions are necessary}). We use the most abstract prompts possible to ensure the approach is generalizable, but we also found that abstract prompts lead to better performance compared to prompts specifically describing how a particular conflict or cultural story is manifested in the climate change debate.

\subsubsection{Results}

Results in \Cref{tab:zero_shot} show that no single model consistently performed best (or worst) across all tasks. Mixtral and Llama are the weakest, especially in stakeholder prediction for for \hero and \villain where both models overpredict entities that are stereotypical heroes and villains for this topic. For instance, they select ``environmental activists'' as heroes and ``pollution'' as villain, despite the fact that they rarely occur in these roles in our articles. In a similar way, they overpredict rare narrative frames as ``Carbon fuelled expansion'' which claims that fossil fuels are necessary for the economy, presumably due to an over-reliance on surface cues (e.g., terms like ``fossil fuels'').
%Thus, weaker models tend to overgeneralize their ``knowledge'' about the topic, disregarding the content and intent of articles.

The strongest models, Sonnet and o1, tie in terms of results, though o1 does better in tasks which require a notion of the overall ``gist'' of the text, such as predicting Cultural story and Conflict. Still, o1 (as well as the other models) does not reliably detect narrative frames based on their description (Narrative task) and tends to excessively predict one class. Thus, despite the fact that human annotators had a high agreement on narrative components, none of the models reaches comparably high performance. Models are also unable to reliably differentiate narrative frames based only on their description, consistent with human performance. In section \Cref{sec:component_labels} we examine if the narrative structure helps the models to predict narrative frames better (as it does for human annotators).

We perform experiments to optimize the prompt and help models learn from examples (see \Cref{app:add_exp}), but they do not lead to performance gains, which shows the difficulty of the tasks.

\paragraph{Effect of the number of classes.}
The difficulty of the Narrative frame prediction task is confounded by the number of classes that need to be distinguished (16). To test whether the performance would increase if the model is asked to choose between a smaller number of classes, we select a sample of three frequent, but similar narratives -- ``12 Years to save the planet'', ``We are all going to die'', and ``Gore'' (see \Cref{app:narratives}), and modify the Narrative prompt to include descriptions only of these three classes. However, this  increases performance only minimally (from F1 of 0.258 to 0.270 for GPT-4o) and nowhere near the level for tasks with a comparable number of classes such as Focus and Conflict. Moreover, even for this simplified task there is a tendency to predict one class, and one of the classes is never chosen correctly (\Cref{fig:confusion_matrix} in Appendix).

\subsubsection{Predicting narrative frames with component labels}
\label{sec:component_labels}

In this section, we explore if using our narrative components (such as specifying \hero, \villain, \victim, and \focus) can improve narrative frame classification. For these experiments we use three models (the strongest Sonnet and middle-grade GPT-4o and Gemini\footnote{We exclude o1 due to its high costs and high variability of results between runs.}) in zero-shot mode. For each of the narrative frame definitions we add an informal description of typical stakeholders for \hero, \villain, \victim (as listed in \Cref{app:narratives} for each of the narrative frames) and the \focus role (see examples of modified prompts in \Cref{app:mod_prompts}). %For example, for the ``Victim blaming'' narrative class we add the explanation of characters involved and the focus and action descriptions (separated by $|$ for readability):

%\begin{myquote}
    %Individuals and consumers are ultimately responsible for taking actions to address climate change. $|$ Both villain and victim here are common people. $|$ The focus is on villain and the narrative criticizes them.
%\end{myquote}

%\begin{table}
%\begin{small}
%    \begin{center}
%    \begin{tabular}{l|cc}
%         & Basic prompt & Structured + oracle structure & Structured  \\
%         \toprule
%         Basic prompt, GPTo &  0.485	& 0.219 \\
%         + HVV &  0.570	& 0.241 \\
%         + Focus and action &  - & 0.316 \\\midrule
%         Zero-shot HVV labels &  0.536	& 0.371\\
%         Oracle HVV labels & 0.728	& 0.718 \\\midrule
%         Basic prompt, GPTo1 & 0.471 & 0.348 \\
%         \bottomrule
%    \end{tabular}
%    \end{center}
%    \end{small}
%    \caption{Macro-averaged F1 scores for Narrative and cultural Story classification, GPT4. \textit{+ HVV} refers to adding hero, villain and victim descriptions, \textit{+ Focus and action} is the addition of focus and action descriptions. The next two lines refer to the structured prompt and also zero-shot or oracle HVV labels for each sample. We also provide the results of GPTo1 for comparison. }
    % \label{tab:str_prompt_results}
%\end{table}

%Lines 1--3 in \Cref{tab:str_prompt_results} show that this strategy increases performance by 0.10 for both the (3-class) Story task and the hardest (16-class) Narrative task.
\begin{figure}
    \centering
    \includegraphics[width=1\linewidth]{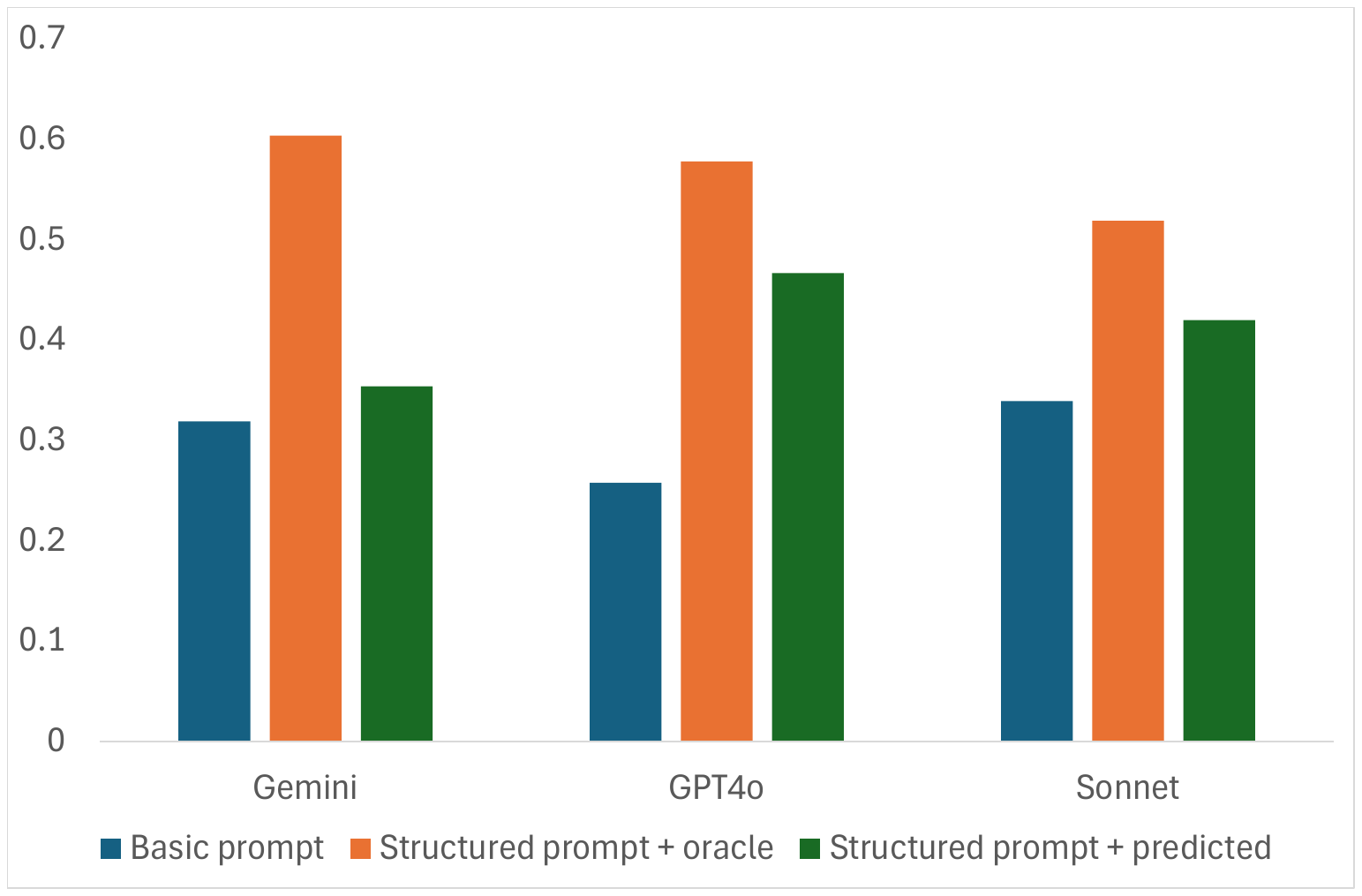}
    \caption{Predicting narrative frames using oracle (human-annotated) and noisy (predicted) labels for their \hero, \villain, \victim, and \focus; the results are macro-averaged F1.}
    \label{fig:struct_pred}
\end{figure}
Next, for each input article we add labels that denote \hero, \villain, and \victim stakeholder categories, as well as \focus, to explicitly represent the structure of the narrative frame. As shown in \Cref{fig:struct_pred} (orange), with {\it oracle} (manually-annotated) labels the performance improves substantially across models (most notably in GPT4o). 
We repeat this experiment with noisy labels predicted by the model, and again observe gains compared to the prompt without structure (green in \Cref{fig:struct_pred}). The models' behavior, however, is quite different: For Gemini we observe minimal improvement, while Sonnet benefits least from the structure labels overall, but does accommodate for the noisiness. GPT4o, despite being the weakest on this task, benefits most from structure and noisy labels, achieving substantially higher performance than the best-performing model (Sonnet) in this condition. Taken together, this shows that explicit structure -- if of reasonably high quality\footnote{These gains occur only when the labels are accurate enough: we observed only minimal gains or drops in performance when using predictions from less strong models.} -- is a more reliable cue for predicting the narrative than its description. 

To examine how introduction of structure affects the narrative frame prediction, and analyze which narrative frames are hard for models to predict even when they are given correct labels for their characters, we compare confusion matrices for Narrative classification with a basic prompt (\Cref{fig:cm_old} in the appendix), and with a structured prompt and oracle character labels (\Cref{fig:cm_new}).
% (\Cref{app:confusion}). 
We observe that before the introduction of structure the predictions are scattered across the matrix; i.e.~both the predictions and errors are not systematic. With the structure, however, we see clear patterns of consolidation: first, narrative frames that have a unique structure (such as ``Officials declare emergency'' which uniquely frames government as a \hero) are now predicted (near) perfectly. Second, errors are now due to confusion of a handful of structurally similar narrative frames, most prevalently two frames that both focus on criticizing the government (\villain), but have different cultural stories: ``12 years to save the Earth'' calls for even more governmental control (\textit{hierarchical}), while ``All talk no action'' opposes government actions (\textit{egalitarian}). This example highlights how cultural stories complement the character role component in our framework, and the importance of each component for effectively differentiating between narrative frames. Future work should explore incorporating all components (and not only characters and focus) in structured prompts.
% and opens the possibility of further improvement by incorporating them.

%Finally, we inspect the narratives that the model with the oracle prompt failed to classify, to determine which aspects apart from characters and focus can be at play. Interestingly, the vast majority of errors was due to confusion between ``All talk no action'' and ``12 years to save the earth''. Both focus on vilifying politicians who fail to support climate goals; however, while in the first narrative the politicians \textit{deny} the necessity to implement measures, in the second they promise to \textit{support} them. Thus, these narratives differ in the villain's action towards the climate policies, which shows that it is also important to consider that aspect of structure.

\section{Narrative Framing of COVID-19}
\label{sec:unsupervised}

In this section we apply our narrative frame structures taxonomy to texts with a different topic and style -- politicians' speeches around the onset of COVID-19 -- to demonstrate its generalizability to other domains. We show how models and prompts developed within the supervised approach (\Cref{sec:supervised}) can be applied to analyze narrative frame components in an unsupervised way.

\subsection{Dataset and model}

We collect transcripts of head-of-state addresses regarding the onset of COVID-19 dating from February to end of July 2020, for three countries: Germany (Angela Merkel; N=12), UK (Boris Johnston; N=24) and Australia (Scott Morrisson; N=6).\footnote{Sources: \url{https://www.bundesregierung.de/breg-en/service/archive/} (official translation into English), \url{https://www.gov.uk/government/speeches/}, \url{https://www.pm.gov.au/media}.} We examine all addresses published during that period and select those that were dedicated to COVID-19.

We use the most reliable model identified in \Cref{sec:supervised} (Claude Sonnet 3.5) in a zero-shot setting. Since the prompts for \focus, \conflict, and \story developed in \Cref{sec:supervised} are domain agnostic, we apply them without changes, only substituting the topic name for ``COVID-19''. However, since the set of stakeholders is likely to be different for this topic, we modify the HVV prompts by replacing the classes with a list of topic-specific stakeholders. We compile this list automatically by generating them from the speeches: first we ask the LLM to extract and merge entities which are likely to represent \hero, \villain, and \victim, then combine the extracted candidates from all speeches and cluster them into groups (prompts in \Cref{app:stakeholder_generation}).

We arrive at a set of 8 stakeholders, some of which are generic and shared with the climate change domain (\textit{government, general public}), while the majority are unique and topic-specific (\textit{vulnerable population, healthcare}, etc). The final set of stakeholders corresponds to prominent stakeholders previously identified as \hero, \villain, \victim in studies on narrative framing in these speeches \citep{bernard2021analysis,mintrom2021policy}.

\subsection{Results}

We apply our approach to discover differences and commonalities in framing of politicians speeches regarding COVID-19. 

First, all speeches across all three politicians were identified as \textit{hero-focused} and \textit{promoting resolution}, which is not surprising given the fact that they are all mobilizing narratives that suggest specific actions to solve the crisis and praise the role of heroes. Similarly, the \villain is consistently detected as ``pandemics", and \victim is ``general public'', especially ``vulnerable populations'', and, later in the period, ``economy''.

However, the stakeholders that are pinpointed as \hero differ across politicians: while all of them recognize the role of ``healthcare workers'', Merkel's speeches also highlight the role of ``general public'', and, later in the pandemic, of ``global efforts''. On the other hand, Morrison's speeches heavily revolve around the role of ``government'' as a hero, as well as ``science experts''. This divergence is in line with prior theoretical analyses of these speeches, which assert that chancellor Merkel recognized the value of combined efforts of the German public and countries around the globe~\citep{mintrom2021policy}, while prime minister Morrison often used reassuring framing relying on the role of science in pandemic management and the imagery of Australia as a ``lucky country''~\citep{bernard2021analysis}.

Similarly, the analysis of predicted cultural stories reveals that Morrison predominantly used \textit{hierarchical} cultural stories (\textit{`Government and following social prescriptions plays the biggest role in managing the crisis}'), Merkel had a larger proportion of \textit{egalitarian} narrative frames than others (\textit{`We must act as one to combat the crisis'}), while Johnson was the only one who alluded to \textit{individualistic} cultural story (\textit{`Take care of yourself and your family'}). Again, these insights align with previous theoretical analyses~\citep{mintrom2021policy}.

% In sum, we showed that LLMs can (to an extent) identify core components of narrative frames when evaluated against human-labelled data, and that our framework generalizes across topics and to scenarios of discovering frame structures in a previously unseen domain.

\section{Conclusion}
We presented a rigorous formalization and taxonomy of components of narrative framing, synthesizing the NPF and \citet{entman1993framing}'s components of a frame. Our method allows to inductively detect narratives from political texts in terms of their character roles, focus, conflict, and underlying cultural story. A high-quality data set of 100 manually labeled articles serves as a benchmark and basis for future annotation bootstrapping. We showed that our framework results in promising improvement of automatic narrative prediction with LLMs, laying a foundation for the important research agenda of large-scale studies of the manifestation and effects of narrative frames. Moreover, we showed that our framework is generalizable to other topics and can assist in exploratory framing analysis without requiring a labeled dataset.

\section{Limitations}
We acknowledge the small size of our data set relative to NLP benchmarks, but emphasize the difficulty of annotating news articles at this level. We prioritize depth over breadth, and our data set can serve both as a benchmark and a high-quality starting point for bootstrapping other story annotations.

Because our approach is inductive / bottom-up we cannot guarantee that the narratives we found cover all possible active narratives or reflect the true narrative distribution. However, since our inductive narratives overlapped with a large part of narratives described in the literature, we are confident that they are representative and comprehensive.

Additional LLM experiments, with larger example pools or advanced reasoning techniques may lead to further improvements but are outside the scope of this work. We showed that incorporating narrative structure into prompts improves performance more substantially than models with advanced reasoning abilities. Future work, however, may want to combine it with such models and techniques.

\section*{Ethics statement}

This study was approved by the University of Melbourne ethics board (Human Ethics Committee LNR 3B), Reference Number 2023-22109-37029- 4, and data acquisition and analysis has been taken out to the according ethical standards. The annotators were compensated at a rate of 35 USD per hour which is well above minimum hourly payment in Australia.

\section*{Acknowledgements}

This work was supported by the ARC Discovery Early Career Research Award (Grant No. DE230100761). We appreciate the computational resources provided for this research by The University of Melbourne’s Research Computing Services and the Petascale Campus Initiative.
We gratefully acknowledge the work of our annotators and the helpful feedback of the anonymous reviewers.

\bibliography{acl_latex}

\begin{thebibliography}{57}
\expandafter\ifx\csname natexlab\endcsname\relax\def\natexlab#1{#1}\fi

\bibitem[{Agarwal and Rambow(2010)}]{agarwal2010automatic}
Apoorv Agarwal and Owen Rambow. 2010.
\newblock Automatic detection and classification of social events.
\newblock In \emph{Proceedings of the 2010 Conference on Empirical Methods in Natural Language Processing}, pages 1024--1034.

\bibitem[{Ali and Hassan(2022)}]{ali2022survey}
Mohammad Ali and Naeemul Hassan. 2022.
\newblock A survey of computational framing analysis approaches.
\newblock In \emph{Proceedings of the 2022 Conference on Empirical Methods in Natural Language Processing}, pages 9335--9348.

\bibitem[{Bamman et~al.(2014)Bamman, Underwood, and Smith}]{bamman-etal-2014-bayesian}
David Bamman, Ted Underwood, and Noah~A. Smith. 2014.
\newblock \href {https://doi.org/10.3115/v1/P14-1035} {A {B}ayesian mixed effects model of literary character}.
\newblock In \emph{Proceedings of the 52nd Annual Meeting of the Association for Computational Linguistics (Volume 1: Long Papers)}, pages 370--379, Baltimore, Maryland. Association for Computational Linguistics.

\bibitem[{Bennett and Edelman(1985)}]{bennett1985toward}
W~Lance Bennett and Murray Edelman. 1985.
\newblock Toward a new political narrative.
\newblock \emph{Journal of communication}.

\bibitem[{Bernard et~al.(2021)Bernard, Basit, Sofija, Phung, Lee, Rutherford, Sebar, Harris, Phung, and Wiseman}]{bernard2021analysis}
Natalie~Reyes Bernard, Abdul Basit, Ernesta Sofija, Hai Phung, Jessica Lee, Shannon Rutherford, Bernadette Sebar, Neil Harris, Dung Phung, and Nicola Wiseman. 2021.
\newblock Analysis of crisis communication by the prime minister of australia during the covid-19 pandemic.
\newblock \emph{International Journal of Disaster Risk Reduction}, 62:102375.

\bibitem[{Bevan(2020)}]{bevan2020climate}
Luke~D Bevan. 2020.
\newblock Climate change strategic narratives in the united kingdom: emergency, extinction, effectiveness.
\newblock \emph{Energy research \& social science}, 69:101580.

\bibitem[{Boswell(2013)}]{boswell2013and}
John Boswell. 2013.
\newblock Why and how narrative matters in deliberative systems.
\newblock \emph{Political studies}, 61(3):620--636.

\bibitem[{Bruner(1991)}]{bruner1991narrative}
Jerome Bruner. 1991.
\newblock The narrative construction of reality.
\newblock \emph{Critical inquiry}, 18(1):1--21.

\bibitem[{Bushell et~al.(2017)Bushell, Buisson, Workman, and Colley}]{bushell2017strategic}
Simon Bushell, G{\'e}raldine~Satre Buisson, Mark Workman, and Thomas Colley. 2017.
\newblock Strategic narratives in climate change: Towards a unifying narrative to address the action gap on climate change.
\newblock \emph{Energy Research \& Social Science}, 28:39--49.

\bibitem[{Chambers and Jurafsky(2009)}]{chambers-jurafsky-2009-unsupervised}
Nathanael Chambers and Dan Jurafsky. 2009.
\newblock \href {https://aclanthology.org/P09-1068} {Unsupervised learning of narrative schemas and their participants}.
\newblock In \emph{Proceedings of the Joint Conference of the 47th Annual Meeting of the {ACL} and the 4th International Joint Conference on Natural Language Processing of the {AFNLP}}, pages 602--610, Suntec, Singapore. Association for Computational Linguistics.

\bibitem[{Crow and Lawlor(2016)}]{crow2016media}
Deserai~A Crow and Andrea Lawlor. 2016.
\newblock Media in the policy process: Using framing and narratives to understand policy influences.
\newblock \emph{Review of Policy Research}, 33(5):472--491.

\bibitem[{Daniels and Endfield(2009)}]{daniels2009narratives}
Stephen Daniels and Georgina~H Endfield. 2009.
\newblock Narratives of climate change: introduction.
\newblock \emph{Journal of Historical Geography}, 35(2):215--222.

\bibitem[{Das et~al.(2024)Das, Chandra, Lee, and Pacheco}]{das-etal-2024-media}
Rohan Das, Aditya Chandra, I-Ta Lee, and Maria~Leonor Pacheco. 2024.
\newblock \href {https://doi.org/10.18653/v1/2024.wnu-1.15} {Media framing through the lens of event-centric narratives}.
\newblock In \emph{Proceedings of the 6th Workshop on Narrative Understanding}, pages 85--98, Miami, Florida, USA. Association for Computational Linguistics.

\bibitem[{Douglas(2007)}]{douglas2007history}
Mary Douglas. 2007.
\newblock A history of grid and group cultural theory.
\newblock \emph{Toronto, Canada: University of Toronto}.

\bibitem[{Entman(1993)}]{entman1993framing}
Robert~M Entman. 1993.
\newblock Framing: Toward clarification of a fractured paradigm.
\newblock \emph{Journal of communication}, 43(4):51--58.

\bibitem[{Finlayson(2012)}]{finlayson2012learning}
Mark Mark~Alan Finlayson. 2012.
\newblock \emph{Learning narrative structure from annotated folktales}.
\newblock Ph.D. thesis, Massachusetts Institute of Technology.

\bibitem[{Fisher(1984)}]{fisher1984narration}
Walter~R Fisher. 1984.
\newblock Narration as a human communication paradigm: The case of public moral argument.
\newblock \emph{Communications Monographs}, 51(1):1--22.

\bibitem[{Fl{\o}ttum and Gjerstad(2017)}]{flottum2017narratives}
Kjersti Fl{\o}ttum and {\O}yvind Gjerstad. 2017.
\newblock Narratives in climate change discourse.
\newblock \emph{Wiley Interdisciplinary Reviews: Climate Change}, 8(1):e429.

\bibitem[{Frermann et~al.(2023)Frermann, Li, Khanehzar, and Mikolajczak}]{frermann-etal-2023-conflicts}
Lea Frermann, Jiatong Li, Shima Khanehzar, and Gosia Mikolajczak. 2023.
\newblock Conflicts, villains, resolutions: Towards models of narrative media framing.
\newblock In \emph{Proceedings of the 61st Annual Meeting of the Association for Computational Linguistics (Volume 1: Long Papers)}, pages 8712--8732. Association for Computational Linguistics.

\bibitem[{Frye(1957)}]{northrop1957anatomy}
Northrop~H Frye. 1957.
\newblock \emph{Anatomy of criticism}.
\newblock Princeton.

\bibitem[{Gehring and Grigoletto(2023)}]{gehring2023analyzing}
Kai~Sebastian Gehring and Matteo Grigoletto. 2023.
\newblock Analyzing climate change policy narratives with the character-role narrative framework.
\newblock \emph{SSRN Electronic Journal}, 10429.

\bibitem[{Greimas(1987)}]{greimas1987meaning}
Algirdas~Julien Greimas. 1987.
\newblock On meaning: Selected writings in semiotic theory.
\newblock \emph{Actants, Actors, and Figures}.

\bibitem[{G{\"u}ss and Tuason(2021)}]{guss2021individualism}
C~Dominik G{\"u}ss and Ma~Teresa Tuason. 2021.
\newblock Individualism and egalitarianism can kill: how cultural values predict coronavirus deaths across the globe.
\newblock \emph{Frontiers in Psychology}, 12:620490.

\bibitem[{Han et~al.(2019)Han, Choi, and Tan}]{han-etal-2019-permanent}
Xiaochuang Han, Eunsol Choi, and Chenhao Tan. 2019.
\newblock \href {https://doi.org/10.18653/v1/N19-1167} {No permanent {F}riends or enemies: Tracking relationships between nations from news}.
\newblock In \emph{Proceedings of the 2019 Conference of the North {A}merican Chapter of the Association for Computational Linguistics: Human Language Technologies, Volume 1 (Long and Short Papers)}, pages 1660--1676, Minneapolis, Minnesota. Association for Computational Linguistics.

\bibitem[{Hu et~al.(2021)Hu, Shen, Wallis, Allen-Zhu, Li, Wang, Wang, and Chen}]{hu2021lora}
Edward~J Hu, Yelong Shen, Phillip Wallis, Zeyuan Allen-Zhu, Yuanzhi Li, Shean Wang, Lu~Wang, and Weizhu Chen. 2021.
\newblock Lo{RA}: Low-rank adaptation of large language models.
\newblock \emph{arXiv preprint arXiv:2106.09685}.

\bibitem[{Iyyer et~al.(2016)Iyyer, Guha, Chaturvedi, Boyd-Graber, and Daum{\'e}~III}]{iyyer-etal-2016-feuding}
Mohit Iyyer, Anupam Guha, Snigdha Chaturvedi, Jordan Boyd-Graber, and Hal Daum{\'e}~III. 2016.
\newblock \href {https://doi.org/10.18653/v1/N16-1180} {Feuding families and former {F}riends: Unsupervised learning for dynamic fictional relationships}.
\newblock In \emph{Proceedings of the 2016 Conference of the North {A}merican Chapter of the Association for Computational Linguistics: Human Language Technologies}, pages 1534--1544, San Diego, California. Association for Computational Linguistics.

\bibitem[{Jahan and Finlayson(2019)}]{jahan-finlayson-2019-character}
Labiba Jahan and Mark Finlayson. 2019.
\newblock \href {https://doi.org/10.18653/v1/W19-2402} {Character identification refined: A proposal}.
\newblock In \emph{Proceedings of the First Workshop on Narrative Understanding}, pages 12--18, Minneapolis, Minnesota. Association for Computational Linguistics.

\bibitem[{Jones(2014)}]{jones2014cultural}
Michael~D Jones. 2014.
\newblock Cultural characters and climate change: How heroes shape our perception of climate science.
\newblock \emph{Social Science Quarterly}, 95(1):1--39.

\bibitem[{Jones et~al.(2023)Jones, Smith-Walter, McBeth, and Shanahan}]{jones2023narrative}
Michael~D Jones, Aaron Smith-Walter, Mark~K McBeth, and Elizabeth~A Shanahan. 2023.
\newblock The narrative policy framework.
\newblock In \emph{Theories of the policy process}, pages 161--195. Routledge.

\bibitem[{Khattab et~al.(2023)Khattab, Singhvi, Maheshwari, Zhang, Santhanam, Vardhamanan, Haq, Sharma, Joshi, Moazam et~al.}]{khattab2023dspy}
Omar Khattab, Arnav Singhvi, Paridhi Maheshwari, Zhiyuan Zhang, Keshav Santhanam, Sri Vardhamanan, Saiful Haq, Ashutosh Sharma, Thomas~T Joshi, Hanna Moazam, et~al. 2023.
\newblock Dspy: Compiling declarative language model calls into self-improving pipelines.
\newblock \emph{arXiv preprint arXiv:2310.03714}.

\bibitem[{Lamb et~al.(2020)Lamb, Mattioli, Levi, Roberts, Capstick, Creutzig, Minx, M{\"u}ller-Hansen, Culhane, and Steinberger}]{lamb2020discourses}
William~F Lamb, Giulio Mattioli, Sebastian Levi, J~Timmons Roberts, Stuart Capstick, Felix Creutzig, Jan~C Minx, Finn M{\"u}ller-Hansen, Trevor Culhane, and Julia~K Steinberger. 2020.
\newblock Discourses of climate delay.
\newblock \emph{Global Sustainability}, 3:e17.

\bibitem[{Levi et~al.(2022)Levi, Mor, Sheafer, and Shenhav}]{levi-etal-2022-detecting}
Effi Levi, Guy Mor, Tamir Sheafer, and Shaul Shenhav. 2022.
\newblock \href {https://doi.org/10.18653/v1/2022.findings-naacl.133} {Detecting narrative elements in informational text}.
\newblock In \emph{Findings of the Association for Computational Linguistics: NAACL 2022}, pages 1755--1765, Seattle, United States. Association for Computational Linguistics.

\bibitem[{Lukin et~al.(2016)Lukin, Bowden, Barackman, and Walker}]{lukin-etal-2016-personabank}
Stephanie Lukin, Kevin Bowden, Casey Barackman, and Marilyn Walker. 2016.
\newblock \href {https://aclanthology.org/L16-1163} {{P}ersona{B}ank: A corpus of personal narratives and their story intention graphs}.
\newblock In \emph{Proceedings of the Tenth International Conference on Language Resources and Evaluation ({LREC}'16)}, pages 1026--1033, Portoro{\v{z}}, Slovenia. European Language Resources Association (ELRA).

\bibitem[{Mintrom et~al.(2021)Mintrom, Rublee, Bonotti, and Zech}]{mintrom2021policy}
Michael Mintrom, Maria~Rost Rublee, Matteo Bonotti, and Steven~T Zech. 2021.
\newblock Policy narratives, localisation, and public justification: responses to covid-19.
\newblock \emph{Journal of European Public Policy}, 28(8):1219--1237.

\bibitem[{Nelson et~al.(1997)Nelson, Oxley, and Clawson}]{nelson1997toward}
Thomas~E Nelson, Zoe~M Oxley, and Rosalee~A Clawson. 1997.
\newblock Toward a psychology of framing effects.
\newblock \emph{Political behavior}, 19:221--246.

\bibitem[{Olsson et~al.(2020)Olsson, Sahlgren, ben Abdesslem, Ekgren, and Eck}]{olsson-etal-2020-text}
Fredrik Olsson, Magnus Sahlgren, Fehmi ben Abdesslem, Ariel Ekgren, and Kristine Eck. 2020.
\newblock \href {https://aclanthology.org/2020.aespen-1.5} {Text categorization for conflict event annotation}.
\newblock In \emph{Proceedings of the Workshop on Automated Extraction of Socio-political Events from News 2020}, pages 19--25, Marseille, France. European Language Resources Association (ELRA).

\bibitem[{Otmakhova et~al.(2024)Otmakhova, Khanehzar, and Frermann}]{otmakhova-etal-2024-media}
Yulia Otmakhova, Shima Khanehzar, and Lea Frermann. 2024.
\newblock \href {https://doi.org/10.18653/v1/2024.acl-long.822} {Media framing: A typology and survey of computational approaches across disciplines}.
\newblock In \emph{Proceedings of the 62nd Annual Meeting of the Association for Computational Linguistics (Volume 1: Long Papers)}, pages 15407--15428, Bangkok, Thailand. Association for Computational Linguistics.

\bibitem[{Patterson and Monroe(1998)}]{patterson1998narrative}
Molly Patterson and Kristen~Renwick Monroe. 1998.
\newblock Narrative in political science.
\newblock \emph{Annual review of political science}, 1(1):315--331.

\bibitem[{Piper et~al.(2021)Piper, So, and Bamman}]{piper-etal-2021-narrative}
Andrew Piper, Richard~Jean So, and David Bamman. 2021.
\newblock \href {https://doi.org/10.18653/v1/2021.emnlp-main.26} {Narrative theory for computational narrative understanding}.
\newblock In \emph{Proceedings of the 2021 Conference on Empirical Methods in Natural Language Processing}, pages 298--311, Online and Punta Cana, Dominican Republic. Association for Computational Linguistics.

\bibitem[{Prince(2003)}]{prince2003dictionary}
Gerald Prince. 2003.
\newblock A dictionary of narratology.
\newblock \emph{U of Nebraska P}.

\bibitem[{Propp(1968)}]{propp1968morphology}
Vladimir Propp. 1968.
\newblock Morphology of the folktale.
\newblock \emph{U of Texas P}.

\bibitem[{Robert and Shenhav(2014)}]{robert2014fundamental}
Dominique Robert and Shaul Shenhav. 2014.
\newblock Fundamental assumptions in narrative analysis: Mapping the field.
\newblock \emph{The qualitative report}, 19(38):1--17.

\bibitem[{Rodrigo-Alsina(2019)}]{rodrigo2019talking}
Miquel Rodrigo-Alsina. 2019.
\newblock Talking about climate change: the power of narratives.
\newblock In \emph{Climate Change Denial and Public Relations}, pages 103--120. Routledge.

\bibitem[{Scheufele and Scheufele(2010)}]{scheufele2010spreading}
Bertram~T Scheufele and Dietram~A Scheufele. 2010.
\newblock Of spreading activation, applicability, and schemas: Conceptual distinctions and their operational implications for measuring frames and framing effects.
\newblock In \emph{{Doing News Framing Analysis}}, pages 126--150. Routledge.

\bibitem[{Semetko and Valkenburg(2000)}]{semetko2000framing}
Holli~A Semetko and Patti~M Valkenburg. 2000.
\newblock Framing european politics: A content analysis of press and television news.
\newblock \emph{Journal of communication}, 50(2):93--109.

\bibitem[{Shahsavari et~al.(2020)Shahsavari, Holur, Wang, Tangherlini, and Roychowdhury}]{shahsavari2020conspiracy}
Shadi Shahsavari, Pavan Holur, Tianyi Wang, Timothy~R Tangherlini, and Vwani Roychowdhury. 2020.
\newblock Conspiracy in the time of corona: automatic detection of emerging covid-19 conspiracy theories in social media and the news.
\newblock \emph{Journal of computational social science}, 3(2):279--317.

\bibitem[{Shanahan et~al.(2018)Shanahan, Jones, and McBeth}]{shanahan2018conduct}
Elizabeth~A Shanahan, Michael~D Jones, and Mark~K McBeth. 2018.
\newblock How to conduct a narrative policy framework study.
\newblock \emph{The Social Science Journal}, 55(3):332--345.

\bibitem[{Shanahan et~al.(2011)Shanahan, McBeth, and Hathaway}]{shanahan2011narrative}
Elizabeth~A Shanahan, Mark~K McBeth, and Paul~L Hathaway. 2011.
\newblock Narrative policy framework: The influence of media policy narratives on public opinion.
\newblock \emph{Politics \& Policy}, 39(3):373--400.

\bibitem[{Shanahan(2007)}]{shanahan2007talking}
Mike Shanahan. 2007.
\newblock Talking about a revolution: climate change and the media.

\bibitem[{Shen et~al.(2023)Shen, Sap, Colon-Hernandez, Park, and Breazeal}]{shen-etal-2023-modeling}
Jocelyn Shen, Maarten Sap, Pedro Colon-Hernandez, Hae Park, and Cynthia Breazeal. 2023.
\newblock \href {https://doi.org/10.18653/v1/2023.emnlp-main.383} {Modeling empathic similarity in personal narratives}.
\newblock In \emph{Proceedings of the 2023 Conference on Empirical Methods in Natural Language Processing}, pages 6237--6252, Singapore. Association for Computational Linguistics.

\bibitem[{Shenhav(2005)}]{shenhav2005thin}
Shaul~R Shenhav. 2005.
\newblock Thin and thick narrative analysis: On the question of defining and analyzing political narratives.
\newblock \emph{Narrative inquiry}, 15(1):75--99.

\bibitem[{Skrynnikova et~al.(2017)Skrynnikova, Astafurova, and Sytina}]{skrynnikova2017power}
IV~Skrynnikova, TN~Astafurova, and NA~Sytina. 2017.
\newblock Power of metaphor: cultural narratives in political persuasion.
\newblock In \emph{7th International Scientific and Practical Conference" Current issues of linguistics and didactics: The interdisciplinary approach in humanities"(CILDIAH 2017)}, pages 279--284. Atlantis Press.

\bibitem[{Sniderman and Theriault(2004)}]{sniderman2004structure}
Paul~M Sniderman and Sean~M Theriault. 2004.
\newblock The structure of political argument and the logic of issue framing.
\newblock \emph{Studies in public opinion: Attitudes, nonattitudes, measurement error, and change}, 3(03):133--65.

\bibitem[{Stammbach et~al.(2022)Stammbach, Antoniak, and Ash}]{stammbach-etal-2022-heroes}
Dominik Stammbach, Maria Antoniak, and Elliott Ash. 2022.
\newblock Heroes, villains, and victims, and {GPT}-3: Automated extraction of character roles without training data.
\newblock In \emph{Proceedings of the 4th Workshop of Narrative Understanding (WNU2022)}, pages 47--56. Association for Computational Linguistics.

\bibitem[{Tangherlini et~al.(2020)Tangherlini, Shahsavari, Shahbazi, Ebrahimzadeh, and Roychowdhury}]{tangherlini2020automated}
Timothy~R Tangherlini, Shadi Shahsavari, Behnam Shahbazi, Ehsan Ebrahimzadeh, and Vwani Roychowdhury. 2020.
\newblock An automated pipeline for the discovery of conspiracy and conspiracy theory narrative frameworks: Bridgegate, pizzagate and storytelling on the web.
\newblock \emph{PloS one}, 15(6):e0233879.

\bibitem[{Thompson(2018)}]{thompson2018cultural}
Michael Thompson. 2018.
\newblock \emph{Cultural theory}.
\newblock Routledge.

\bibitem[{Zhao et~al.(2024)Zhao, Tu, Du, and Xue}]{zhao-etal-2024-media}
Jin Zhao, Jingxuan Tu, Han Du, and Nianwen Xue. 2024.
\newblock \href {https://doi.org/10.18653/v1/2024.emnlp-main.954} {Media attitude detection via framing analysis with events and their relations}.
\newblock In \emph{Proceedings of the 2024 Conference on Empirical Methods in Natural Language Processing}, pages 17197--17210, Miami, Florida, USA. Association for Computational Linguistics.

\end{thebibliography}

\appendix

\section{Statistics for the US Climate articles dataset}
\label{app:dataset_stats}

\begin{figure}[h]
    \centering
    \includegraphics[width=0.75\linewidth]{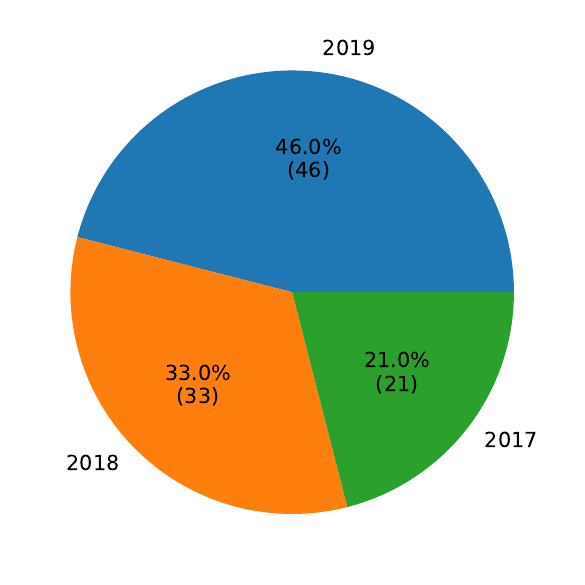}
    \caption{Distribution of articles across years}
    \label{fig:year_stats}
\end{figure}

\begin{figure}[h]
    \centering
    \includegraphics[width=0.9\linewidth]{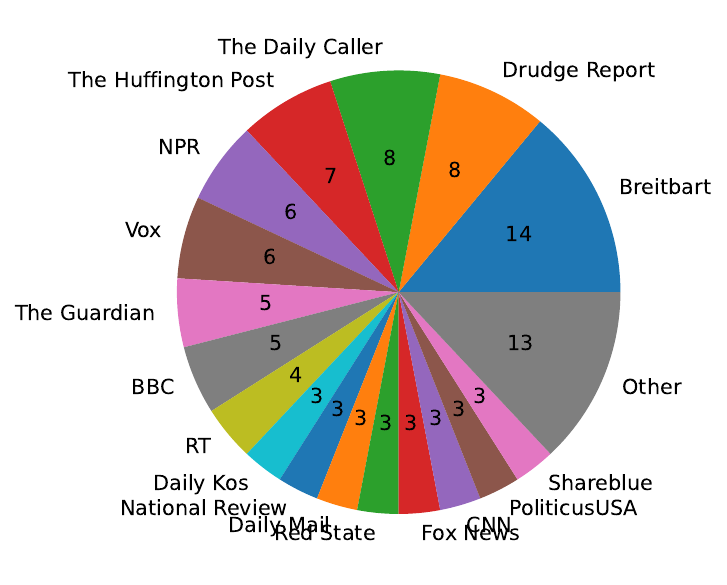}
    \caption{Distribution of articles across media outlets}
    \label{fig:outlet_stats}
\end{figure}

\begin{figure}[h]
    \centering
    \includegraphics[width=0.75\linewidth]{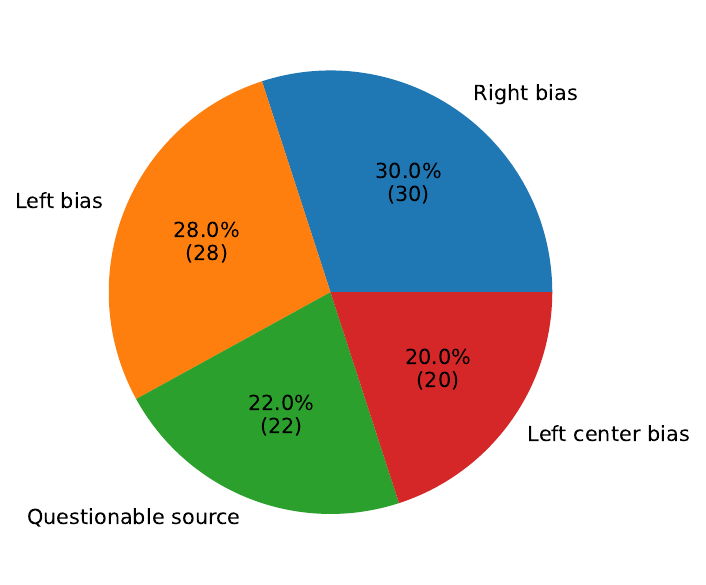}
    \caption{Distribution of articles across political leanings of the outlets}
    \label{fig:leaning_stats}
\end{figure}

Figures \ref{fig:year_stats} to \ref{fig:leaning_stats} show the distribution of articles in the US climate narratives dataset according to their publication year (\Cref{fig:year_stats}), outlet (\Cref{fig:outlet_stats}), and the political leaning of the latter (\Cref{fig:leaning_stats}), as identified by the Media Bias Fact Check (MBFC) website\footnote{\url{https://mediabiasfactcheck.com/}}.

\section{Identifying the main characters}
\label{app:deriving}

As we explain in \Cref{sec:character_annotation}, not all entities in an article represent its main \hero, \villain, or \victim. To be able to reliably and consistently identify the main characters, we adhere to the following process:

(1) We consider only the entities which are consistent with the overall stance of the article. In particular, journalists often cite the opposing view, and thus can mention a set of characters which is different from the one aligned with stance. For example, in the article in \Cref{fig:hvv}, melting ice is a \victim of rising temperatures, according to the viewpoint of climate activists. However, while the author cites this viewpoint, it does not reflect the main message of the article, so the corresponding entities are not considered as potential \hero, \villain, and \victim. To summarize, the main characters are the ones framed so by the author/narrator.

% \begin{enumerate}
(2) We discard characters that either form the backdrop of the story or are used to illustrate a minor (often competing) idea within the main narrative.\footnote{We are well aware that news stories are complex in terms of interplay of narratives within them and most of them contain what \citet{flottum2017narratives} refers to as narrative polyphony. We intentionally restrict the task to identification of the main narrative only as the first step in disentangling narrative complexity.} For example, in the narrative in \Cref{fig:hvv} melting ice or Arctic animals are only included as a point of tension between the climate activists and denialists but don't themselves play a major role in the narrative. The main characters are the parties expressing their opinion regarding them.

%Other examples of this - prob for Appendix for later? scientific reports/legislations can only provide a backdrop for the argument rather than be active parts (Narr 716, ; a story focusing on the flood devastation can briefly mention someone who calls people to action, but it does not become a Officials declare the emergency narrative.
%differentiate between Gore and Alarmist?

(3) The same character can be referred to several times and be represented with several stakeholders. For example, it is common for news stories to mention both climate change regulations/policies and the politicians that propose them. In such cases, instead of adding multiple stakeholders for a character, we choose the one that was most prominent in the context or can be used to infer the other (for example, we would choose policies over politicians if the article focuses on them). 

(4) We only consider characters that are active in the plot, rather than references to potential or past heroes, villains, and victims. For example, a news story that paints Republicans as a \villain for not implementing climate change measures\footnote{Article 512 in our dataset.} concludes with the following sentence:

\begin{myquote}
    For 2020 and beyond, climate justice will have to become the most animating issue for Democrats.
\end{myquote}

Since the positive impact of Democrats is only hoped for or predicted to happen in the future, Democrats are not an active \hero, and, overall, the \hero in this news story is absent. It is important not to assign extraneous entities to the character slots, even if they are otherwise empty, as it will later help to differentiate between narratives. For example, here it allows us to distinguish a narrative criticizing the \villain from alternative narratives which depict an active conflict between \hero and \villain.

(5) For the same reasons, we do not add stakeholders that are only implied but not directly referred to in the text. For example, we do not add ``environment'' as a \victim unless it is specifically mentioned, though it can be inferred from the majority of pro-climate action news stories. Similarly, though the stories warning about the dangers of climate inaction are usually inspired by scientific evidence, scientists or scientific reports are not a \hero in them unless they have an active role, as in here\footnote{Article 537 in our dataset.}

\begin{myquote}
    Climate report warns of extreme weather, displacement of millions without action
\end{myquote}

This allows to differentiate between a narrative which appeals to authority of scientists (so called ``Gore'' narrative) from a similar but often more emotionally charged and less ``objective'' alarmist narrative (``12 Years to save the world'') (see \Cref{app:narratives} for detailed description).

\section{Stakeholder categories}
\label{app:stakeholder}

We use the following 10 stakeholder categories from \citep{frermann-etal-2023-conflicts}:

\noindent
GOVERNMENTS\_POLITICIANS: governments and political organizations

\noindent
INDUSTRY\_EMISSIONS: industries, businesses, and the pollution created by them

\noindent
LEGISLATION\_POLICIES: policies and legislation responses

\noindent
GENERAL\_PUBLIC: general public, individuals, and society, including their wellbeing, status quo and economy

\noindent
ANIMALS\_NATURE\_ENVIRONMENT: nature and environment in general or specific species

\noindent
ENV.ORGS\_ACTIVISTS: climate activists and organizations

\noindent
SCIENCE\_EXPERTS\_SCI.REPORTS: scientists and scientific reports/research

\noindent
CLIMATE\_CHANGE: climate change as a process or consequence

\noindent
GREEN\_TECHNOLOGY\_INNOVATION: innovative and green technologies

\noindent
MEDIA\_JOURNALISTS: media and journalists

%\section{Action types}
%\label{sec:action_types}

%We use the following set of verbs to classify actions:
%\\
%ABANDON: the story describes abandoning climate change policies\\
%CALL\_TO\_ACTION: the story contains a call to action regarding climate change\\
%CHANGE\_BEHAVIOUR: The story describes changing behaviour to address climate\\ change\\
%CRITICIZE: the story criticizes climate policies or someone's actions regarding it\\
%DENY: the story denies the necessity of climate change policies\\
%ENDANGER: the story describes endangering someone because of climate change\\
%ENFORCE: the story describes enforcing climate change policies\\
%EXPOSE: the story exposes flaws of climate change policies or people advocating for them\\
%INTRODUCE: the story describes introducing measures to address climate change\\
%POLITICAL\_ACTION: the story describes political actions regarding climate change such as protests or legal actions\\
%PROPOSE: the story describes the proposal of climate change policies\\
%PROTECT: the story describes protection from climate change\\
%RAISE\_ALARM: the story raises alarm regarding climate change\\
%RENEGE: the story describes reneging on promises regarding climate change policies\\
%THWART: the story describes thwarting climate change measures

\section{Annotation process}
\label{app:annotation}
The annotation was performed in two stages:

\subsection{Stage 1: Annotating \hero, \villain, and \villain}

\begin{figure*}
    \centering
    \includegraphics[width=0.9\linewidth]{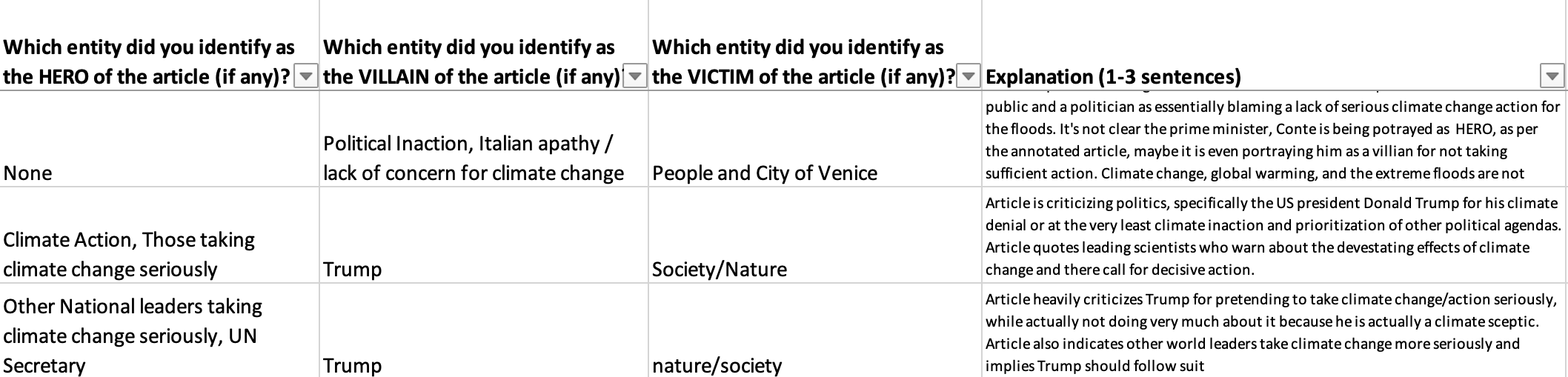}
    \caption{Example of stage 1 annotations (\hero, \villain, \victim)}
    \label{fig:ann1_example}
\end{figure*}

During stage 1, we employed three external annotators, all with an academic background in the social sciences and familiar with the Narrative Policy Framework, and one of the authors of the article, who is considered an expert annotator with knowledge of media discourse and framing. For each article, each annotator had to (1) read it beginning to end and (2) identify the main \hero, \villain and \victim (if any) and record them in free form based on the procedure described in \Cref{app:deriving}. They were also asked to record their reasoning in plain text (see an example and annotation interface in \Cref{fig:ann1_example}).

%The external annotators were expected to spend around 10 minutes per article, and were compensated with 35 USD per hour.
Each of the 100 articles in the dataset was annotated at least by two external annotators, and all of them were in addition annotated by the internal expert annotator. Since the annotators were asked to specify entities for \hero, \villain, and \victim in free form, their annotations were not directly comparable (e.g. "Biden" vs "Joe Biden" vs "Democrats"). Thus, to evaluate the annotation agreement, as well as to convert the data to a more abstract and useful structure (see \Cref{sec:character_annotation}), the expert annotator mapped the specific characters mentioned by each of the annotators to their stakeholder classes (\Cref{app:stakeholder}). 

We evaluate the agreement between all four annotators using Krippendorf's $\alpha$, and report the averaged agreement of each of the three external annotators with the expert. For the latter, we use the standard metrics of agreement rate (=accuracy), Cohen's $\kappa$, and the less commonly used Gwet's AC1, which compensates for the high imbalance in data distribution. The resulting inter-annotator agreement statistics can be found in \Cref{tab:iaa_stage1}. Overall, we observe acceptable to strong levels of agreement between all four annotators, as well as very high average agreement of each of the annotators with the expert (as judged based on Gwet's AC1). A relatively lower agreement for \hero and \villain in comparison to \victim is explained by the fact that the annotators sometimes chose entities belonging to different stakeholder types to represent the same event. For example, a particular climate initiative can be represented both by a parliamentary bill such as New Green Deal (LEGISLATION\_POLICIES), and by the group of people behind it (GOVERNMENTS\_POLITICIANS). In case of disagreement the final label was chosen based on majority vote.

\begin{table}
\begin{small}
    \begin{center}
    \begin{tabular}{l|ccc}
         & Hero & Villain & Victim \\
         \toprule
         Krippendorff's $\alpha$ &  0.757 &  0.673 &  0.812 \\
         Agreement rate & 0.852 & 0.855 & 0.927 \\
         Cohen's $\kappa$ &  0.783 & 0.745 & 0.876 \\
         Gwet't AC1 & 0.837 & 0.843 & 0.914\\
         \bottomrule
    \end{tabular}
    \end{center}
    \end{small}
    \caption{Inter-annotator agreement for \hero, \villain, and \victim annotation}
    \label{tab:iaa_stage1}
\end{table}

\subsection{Stage 2: Annotating \focus, \conflict, and \story}

\begin{figure*}
    \centering
    \includegraphics[width=1\linewidth]{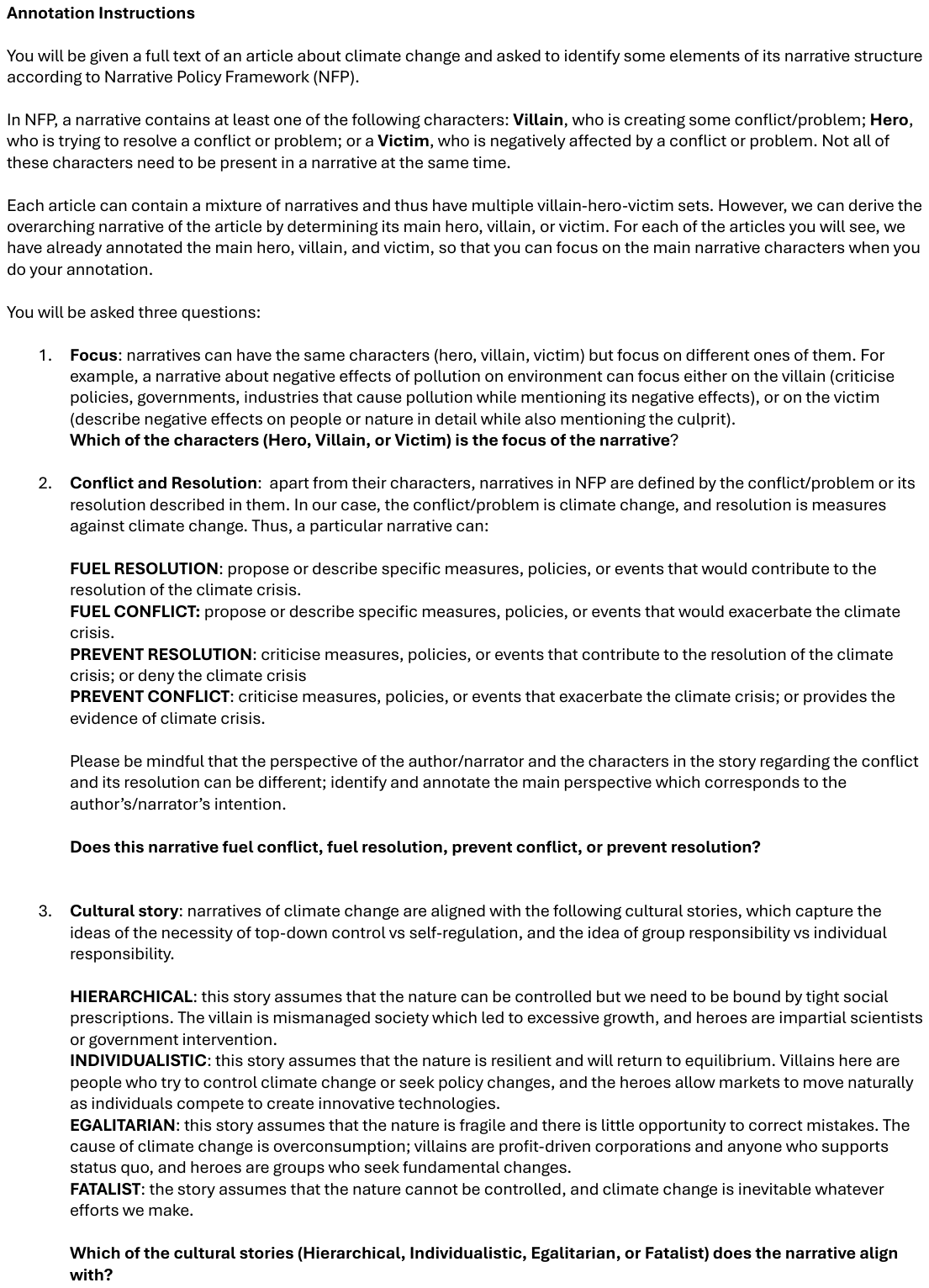}
    \caption{Instructions for Stage 2 annotation (\focus, \conflict, \story)}
    \label{fig:stage2_instructions}
\end{figure*}

\begin{figure*}
    \centering
    \includegraphics[width=0.7\linewidth]{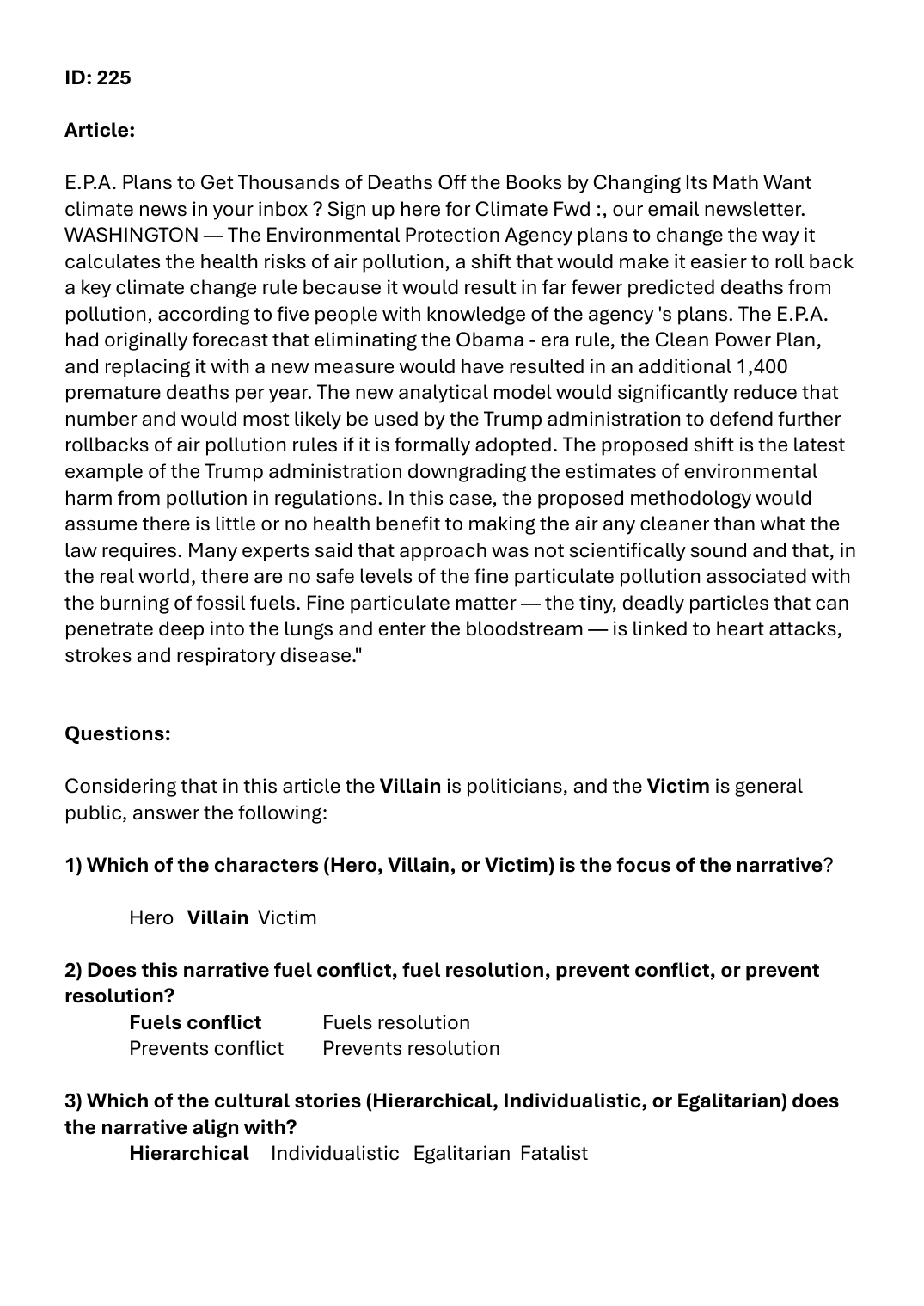}
    \caption{An example of Stage 2 annotation (\focus, \conflict, \story)}
    \label{fig:stage2_ann}
\end{figure*}

In the second stage, the expert annotator annotated all 100 articles in terms of their \focus, \conflict, and \story. Next, a random sample of 30 articles was annotated by another internal annotator who is also an expert in Narrative Policy Framework and framing analysis. The instructions for the annotation and an example of an annotated article are shown in \Cref{fig:stage2_instructions,fig:stage2_ann} respectively. To ensure a high quality of the resulting dataset, all disagreements were discussed and adjudicated, and then the corresponding changes were reflected in the samples beyond this calibration study, if necessary. 

\Cref{tab:iaa_stage2} shows the agreement statistics between the two annotators in Stage 2, using the same metrics as for Stage 1. We observe high agreement rates for all three classes, with other scores varying slightly due to number of classes and class distribution, but all being within the strong or very strong agreement range. Disagreement analysis revealed that there were disagreements on \focus (between \villain and \victim), when both were discussed at similar length and depth in the article. For \conflict and Cultural story, the disagreements were more systematic (such as confusion between Fuel Resolution and Prevent Conflict, or between Hierarchical and Egalitarian stories); the insights arising from the discussion were reflected in the final labels and allowed us to refine the definitions of these concepts for the prompts used in LLM experiments. 

\begin{table}
\begin{small}
    \begin{center}
    \begin{tabular}{l|ccc}
         & Focus & Conflict & Cultural story \\
         \toprule
         Krippendorf's $\alpha$ & 0.780 &  0.820 & 0.801 \\
         Agreement rate & 0.867 & 0.867 &  0.867 \\
         Cohen's $\kappa$ & 0.776 &  0.817 &  0.800\\
         Gwet't AC1 & 0.810 & 0.824 & 0.801 \\
         \bottomrule
    \end{tabular}
    \end{center}
    \end{small}
    \caption{Inter-annotator agreement for \focus, \conflict, and \story annotation}
    \label{tab:iaa_stage2}
\end{table}

% To ensure a high quality dataset, 

\subsection*{Annotation with vs without narrative frame structure}
\label{app:str_annotation}

We empirically tested if structural components help to differentiate between narratives in human annotation. Specifically, we compare agreement in narrative detection when using a structure-based annotation approach (bottom-up; as described above) vs using a more traditional approach where the annotators are asked to classify narratives top-down based on their descriptions.

For the structure-based approach, we estimate the agreement based on the sample of 30 articles we used for Stage 2 annotation (see \Cref{app:annotation}). In particular, we assume that both annotators agree on a particular narrative if they choose exactly the same values for all its components. For the traditional approach, we ask two annotators who took part in Stage 1 of annotation (and thus did not classify any elements of the narrative except for its characters) to choose a narrative frame for each article based on its description only (as listed in \Cref{app:narratives}). 

We find that annotation using our narrative structures resulted in 63\% agreement, while top-down annotation based on the narrative frame descriptions resulted in a substantially lower 37\%. Thus, we can tentatively conclude that structure-based analysis improves narrative detection and understanding. We also observed a reduction in time required for annotation (15 minutes per article based on descriptions of narrative frames vs 7 minutes per article based on its structure, on average).

\newpage

\section{US Climate Narratives: Structures and description}
\label{app:narratives}

In this section we list the 16 discovered narratives in the US climate change study, their structures, references to the literature where they have been discussed, and exact definitions taken from that source.

\subsection{Narratives focusing on \hero}

\subsubsection{You're destroying our future}

\textbf{Hero}: ENV.ORGS\_ACTIVISTS

\noindent
\textbf{Villain}: GOVERNMENTS\_POLITICIANS

\noindent
\textbf{Victim}: <optional>

\noindent
\textbf{Conflict}: FUEL RESOLUTION

\noindent
\textbf{Cultural story}: EGALITARIAN

\noindent
\textbf{Description}: The political stasis around climate change means that we cannot rely on politicians to create the change necessary. With collective action, even the politically weak can make a difference and secure a future for generations to come. This can manifest as anything from protests (school strikes) to non-violent civil disobedience.

\noindent
\textbf{Source}: \citet{bevan2020climate}

\subsubsection{Technological optimism}

\textbf{Hero}: GREEN\_TECHNOLOGY\_INNOVATION

\noindent
\textbf{Villain}: INDUSTRY\_EMISSIONS, CLIMATE\_CHANGE

\noindent
\textbf{Victim}: <optional>

\noindent
\textbf{Conflict}: FUEL RESOLUTION

\noindent
\textbf{Cultural story}: EGALITARIAN
    
\noindent
\textbf{Description}: We should focus our efforts on current and future technologies, which will unlock great possibilities for addressing climate change.

\noindent
\textbf{Source}: \citet{lamb2020discourses}

\subsubsection{Officials declare emergency}

\textbf{Hero}: GOVERNMENTS\_POLITICIANS

\noindent
\textbf{Villain}: INDUSTRY\_EMISSIONS, CLIMATE\_CHANGE, GOVERNMENTS\_POLITICIANS

\noindent
\textbf{Victim}: <optional>

\noindent
\textbf{Conflict}: FUEL RESOLUTION

\noindent
\textbf{Cultural story}: HIERARCHICAL

\noindent
\textbf{Description}: The climate crisis is sufficiently severe that it warrants declaring a climate emergency. This should occur at different levels of government as climate requires action at all levels, from the hyper-local to the global.

\noindent
\textbf{Source}: \citet{bevan2020climate}

\subsubsection{Every little helps}

\textbf{Hero}: GENERAL\_PUBLIC

\noindent
\textbf{Villain}: GENERAL\_PUBLIC

\noindent
\textbf{Victim}: <optional>

\noindent
\textbf{Conflict}: FUEL RESOLUTION

\noindent
\textbf{Cultural story}: INDIVIDUALISTIC

\noindent
\textbf{Description}: This narrative presents a society which has transitioned to a sustainable ``green'' way of life. Could be expressed by portraying individuals as the protagonists of stories that propose solutions to climate change.

\noindent
\textbf{Source}: \citet{bushell2017strategic}

\subsection{Narratives focusing on \villain}

\subsubsection{12 years to save the world}

\textbf{Hero}: <optional>

\noindent
\textbf{Villain}: GOVERNMENTS\_POLITICIANS

\noindent
\textbf{Victim}: ANIMALS\_NATURE\_ENVIRONMENT, GENERAL\_PUBLIC, CLIMATE\_CHANGE

\noindent
\textbf{Conflict}: PREVENT CONFLICT

\noindent
\textbf{Cultural story}: HIERARCHICAL

\noindent
\textbf{Description}: Past and present human action (or inaction) risks a catastrophic future climatic event unless people change their behaviour to mitigate climate change.

\noindent
\textbf{Source}: \citet{bevan2020climate}

\subsubsection{Gore}

\textbf{Hero}: SCIENCE\_EXPERTS\_SCI.REPORTS

\noindent
\textbf{Villain}: GOVERNMENTS\_POLITICIANS, GENERAL\_PUBLIC, INDUSTRY\_EMISSIONS

\noindent
\textbf{Victim}: ANIMALS\_NATURE\_ENVIRONMENT, CLIMATE\_CHANGE

\noindent
\textbf{Conflict}: FUEL RESOLUTION

\noindent
\textbf{Cultural story}: HIERARCHICAL

\noindent
\textbf{Description}: This is a narrative of scientific discovery which climaxes on the certainty that climate change is unequivocally caused by humans.

\noindent
\textbf{Source}: \citet{bushell2017strategic}

\subsubsection{The collapse is imminent}

\textbf{Hero}: ENV.ORGS\_ACTIVISTS

\noindent
\textbf{Villain}: GOVERNMENTS\_POLITICIANS

\noindent
\textbf{Victim}: <optional>

\noindent
\textbf{Conflict}: FUEL RESOLUTION

\noindent
\textbf{Cultural story}: EGALITARIAN

\noindent
\textbf{Description}: The climate crisis is such that some kind of societal collapse is near inevitable. Due to the inaction of the negligent or complacent politicians the social contract has broken down and it is incumbent upon individuals to engage in non-violent civil disobedience to shock society into urgent action.

\noindent
\textbf{Source}: \citet{bevan2020climate}

\subsubsection{Climate solutions won't work}

\textbf{Hero}: <optional>

\noindent
\textbf{Villain}: LEGISLATION\_POLICIES, GREEN\_TECHNOLOGY\_INNOVATION

\noindent
\textbf{Victim}: GENERAL\_PUBLIC, ANIMALS\_NATURE\_ENVIRONMENT

\noindent
\textbf{Conflict}: PREVENT RESOLUTION

\noindent
\textbf{Cultural story}: INDIVIDUALISTIC

\noindent
\textbf{Description}: Climate policies are harmful and a threat to society and the economy. Climate policies are ineffective and too difficult to implement.

\noindent
\textbf{Source}: \citet{lamb2020discourses}

\subsubsection{No sticks just carrots}

\textbf{Hero}: LEGISLATION\_POLICIES

\noindent
\textbf{Villain}: LEGISLATION\_POLICIES

\noindent
\textbf{Victim}: GENERAL\_PUBLIC

\noindent
\textbf{Conflict}: PREVENT RESOLUTION

\noindent
\textbf{Cultural story}: INDIVIDUALISTIC

\noindent
\textbf{Description}: Society will only respond to supportive and voluntary policies, restrictive measures will fail and should be abandoned.

\noindent
\textbf{Source}: \citet{lamb2020discourses}

\subsubsection{All talk little action}

\textbf{Hero}: <optional>

\noindent
\textbf{Villain}: GOVERNMENTS\_POLITICIANS

\noindent
\textbf{Victim}: <optional>

\noindent
\textbf{Conflict}: PREVENT RESOLUTION

\noindent
\textbf{Cultural story}: EGALITARIAN

\noindent
\textbf{Description}: This narrative emphasises inconsistency between ambitious climate action targets and actual actions.

\noindent
\textbf{Source}: \citet{lamb2020discourses}

\subsubsection{Victim blaming}

\textbf{Hero}: <optional>

\noindent
\textbf{Villain}: GENERAL\_PUBLIC

\noindent
\textbf{Victim}: GENERAL\_PUBLIC

\noindent
\textbf{Conflict}: PREVENT RESOLUTION

\noindent
\textbf{Cultural story}: INDIVIDUALISTIC

\noindent
\textbf{Description}: Individuals and consumers are ultimately responsible for taking actions to address climate change.

\noindent
\textbf{Source}: \citet{lamb2020discourses}

\subsubsection{Debate and scam}

\textbf{Hero}: <optional>

\noindent
\textbf{Villain}: GOVERNMENTS\_POLITICIANS, LEGISLATION\_POLICIES, ENV.ORGS\_ACTIVISTS, MEDIA\_JOURNALISTS

\noindent
\textbf{Victim}: <optional>

\noindent
\textbf{Conflict}: PREVENT RESOLUTION

\noindent
\textbf{Cultural story}: INDIVIDUALISTIC

\noindent
\textbf{Description}: The heroes of this narrative are sceptical individuals who dare to challenge the false consensus on climate change which is propagated by those with vested interests.

\noindent
\textbf{Source}: \citet{lamb2020discourses}

\subsubsection{Others are worse than us}

\textbf{Hero}: GOVERNMENTS\_POLITICIANS

\noindent
\textbf{Villain}: GOVERNMENTS\_POLITICIANS

\noindent
\textbf{Victim}: <optional>

\noindent
\textbf{Conflict}: PREVENT RESOLUTION

\noindent
\textbf{Cultural story}: INDIVIDUALISTIC

\noindent
\textbf{Description}: Other countries, cities or industries are worse than ourselves. There is no point for us to  implement climate policies, because we only cause a small fraction of the emissions. As long as others emit even more than us, actions won’t be effective.

\noindent
\textbf{Source}: \citet{lamb2020discourses}

\subsection{Narratives focusing on \victim}

\subsubsection{Endangered species}

\textbf{Hero}: <optional>

\noindent
\textbf{Villain}: GOVERNMENTS\_POLITICIANS, LEGISLATION\_POLICIES, INDUSTRY\_EMISSIONS

\noindent
\textbf{Victim}: ANIMALS\_NATURE\_ENVIRONMENT

\noindent
\textbf{Conflict}: PREVENT CONFLICT

\noindent
\textbf{Cultural story}:HIERARCHICAL

\noindent
\textbf{Description}: Endangered species (like polar bears) are the helpless victims of this narrative, who are seeing their habitat destroyed by the actions of villainous humans.

\noindent
\textbf{Source}: \citet{bushell2017strategic}

\subsubsection{We are all going to die}

\textbf{Hero}: <optional>

\noindent
\textbf{Villain}: CLIMATE\_CHANGE, INDUSTRY\_EMISSIONS

\noindent
\textbf{Victim}: GENERAL\_PUBLIC

\noindent
\textbf{Conflict}: PREVENT CONFLICT

\noindent
\textbf{Cultural story}: EGALITARIAN

\noindent
\textbf{Description}: This narrative shows the current or potential catastrophic impact of climate change on people.

\noindent
\textbf{Source}: \citet{shanahan2007talking}

\subsubsection{Carbon fueled expansion}

\textbf{Hero}: <optional>

\noindent
\textbf{Villain}: LEGISLATION\_POLICIES, GREEN\_TECHNOLOGY\_INNOVATION

\noindent
\textbf{Victim}: GENERAL\_PUBLIC, INDUSTRY\_EMISSIONS

\noindent
\textbf{Conflict}: PREVENT RESOLUTION

\noindent
\textbf{Cultural story}: INDIVIDUALISTIC

\noindent
\textbf{Description}:The free market is at the centre of this narrative which presents action on climate change as an obstacle to the freedom and well-being of citizens. The narrative can stress social justice or well-being of individual citizens.

\noindent
\textbf{Source}: \citet{bushell2017strategic}

\section{Model sizes, costs and parameters}
\label{app:size}

% \noindent mixtral-8x7B-Instruct-v0.1 (Mixtral): 46.7B params\\

% \noindent gemini-1.5-pro (Gemini): 1.5T params\\

% \noindent Llama-3.1-8B-Instruct: 8B params \\

% \noindent Approximate experiments costs: 600 USD.\\

% \noindent Hyperparameters for Llama LoRA fine-tuning:\\

% \noindent max seq length = 4000\\
% \noindent r = 16\\
% \noindent lora alpha = 16\\
% \noindent lora dropout = 0\\
% \noindent learning rate = 2e-4\\
% \noindent optim = adamw8bit\\
% \noindent weight decay = 0.01\\

% \begin{table}[h!]
\centering
\begin{tabular}{@{}ll@{}} % @{} removes extra space at the left/right of the table
\toprule
\multicolumn{2}{@{}l}{\textbf{Model Parameters}} \\
\midrule
Mixtral-8x7B-Instruct-v0.1 & 46.7B params \\
Gemini-1.5-Pro & 1.5T params \\
Llama-3.1-8B-Instruct & 8B params \\
\midrule
\multicolumn{2}{@{}l}{\textbf{Experiment Costs}} \\
\midrule
Approximate costs & 600 USD \\
\midrule
\multicolumn{2}{@{}l}{\textbf{Hyperparameters for Llama LoRA Fine-tuning}} \\
\midrule
Max sequence length & 4000 \\
$r$ (LoRA rank) & 16 \\
LoRA alpha & 16 \\
LoRA dropout & 0 \\
Learning rate & $2 \times 10^{-4}$ \\
Optimizer & adamw8bit \\
Weight decay & 0.01 \\
\bottomrule
\end{tabular}
% \caption{Overview of models and fine-tuning parameters}
% \label{tab:model_training_details}

% \end{table}

\section{Annotated dataset statistics}
\label{app:data_stats}

\begin{figure*}[h!]
    \centering
    \includegraphics[width=1\linewidth]{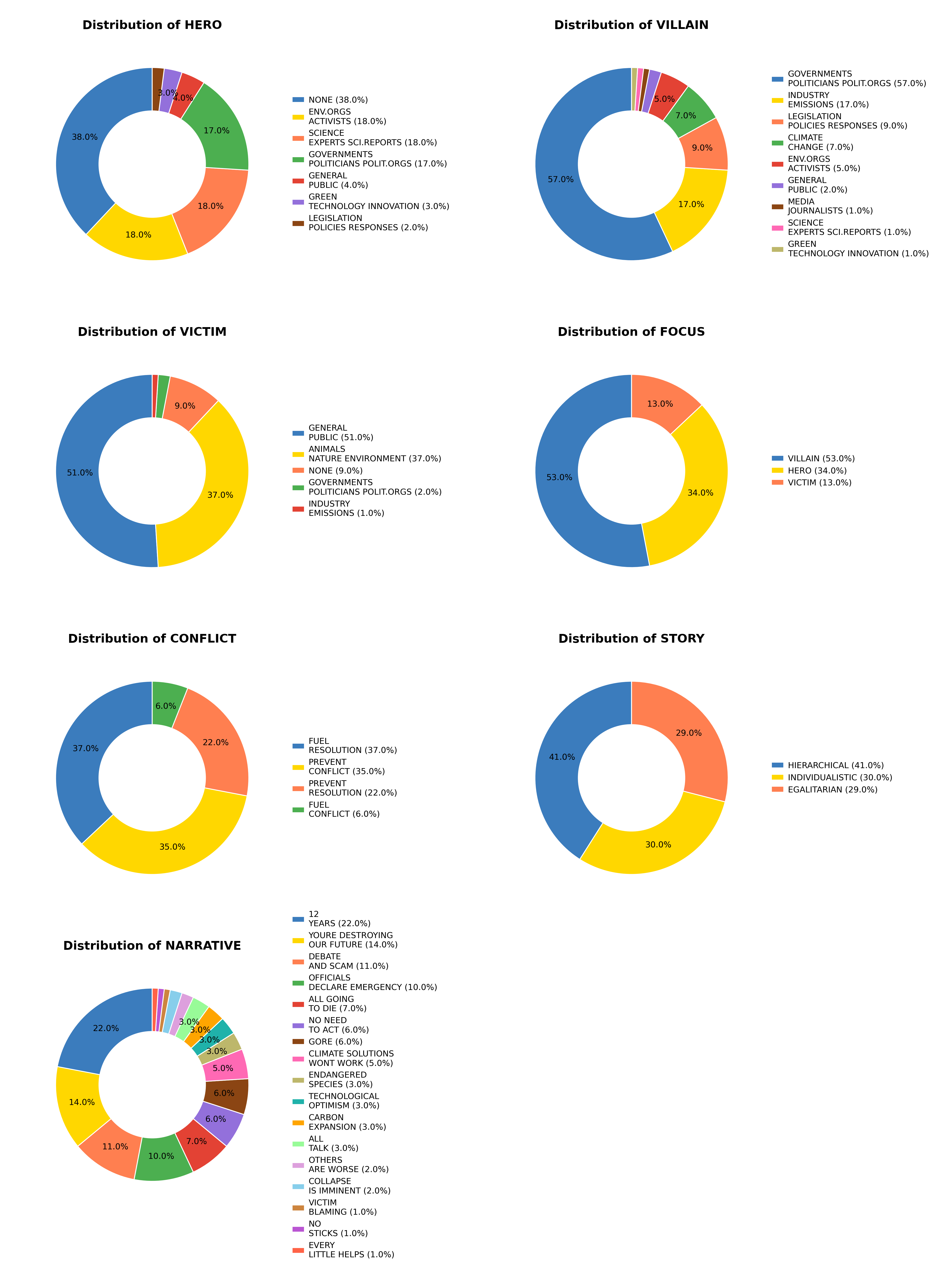}
    \caption{Label distributions for narrative frames and their components in our labelled dataset of 100 US climate change news articles.}
    \label{fig:dist}
\end{figure*}

In \Cref{fig:dist} we show the distribution of all components of our framework (Hero, Villain, Victim stakeholders; Focus; Conflict; Cultural Story), as well as final narratives across the 100 articles. 

\section{Distribution of narrative frame components across political leanings}
\label{app:frame_leanings}

\begin{figure}[h!]
    \centering
    \includegraphics[width=1\linewidth]{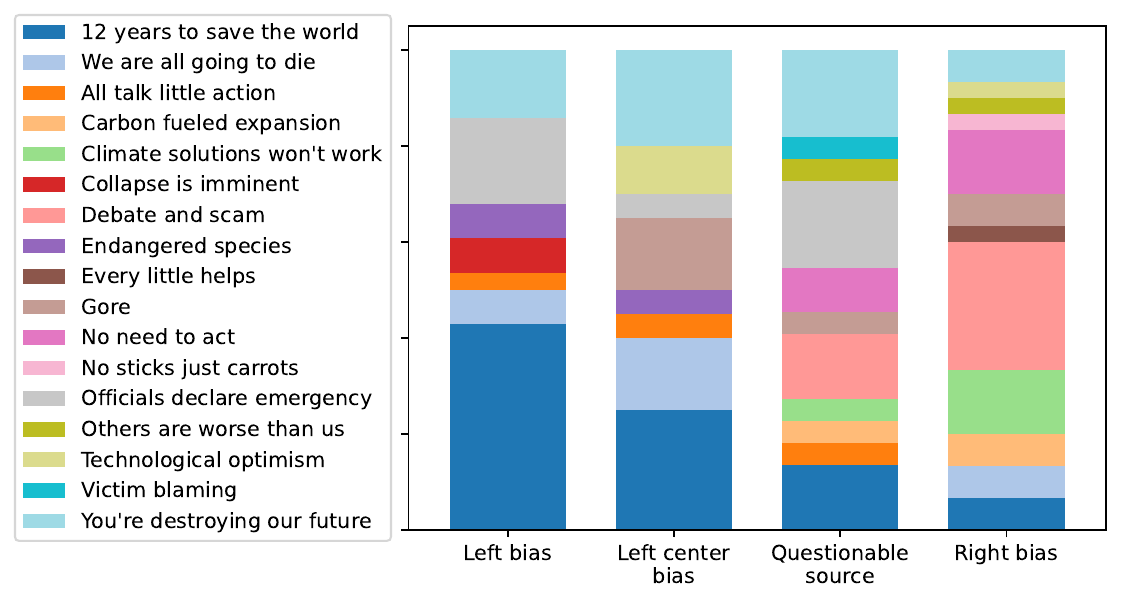}
    \caption{Distribution of narrative frames across political leanings}
    \label{fig:narr_leaning}
\end{figure}

\begin{figure}[h!]
    \centering
    \includegraphics[width=1\linewidth]{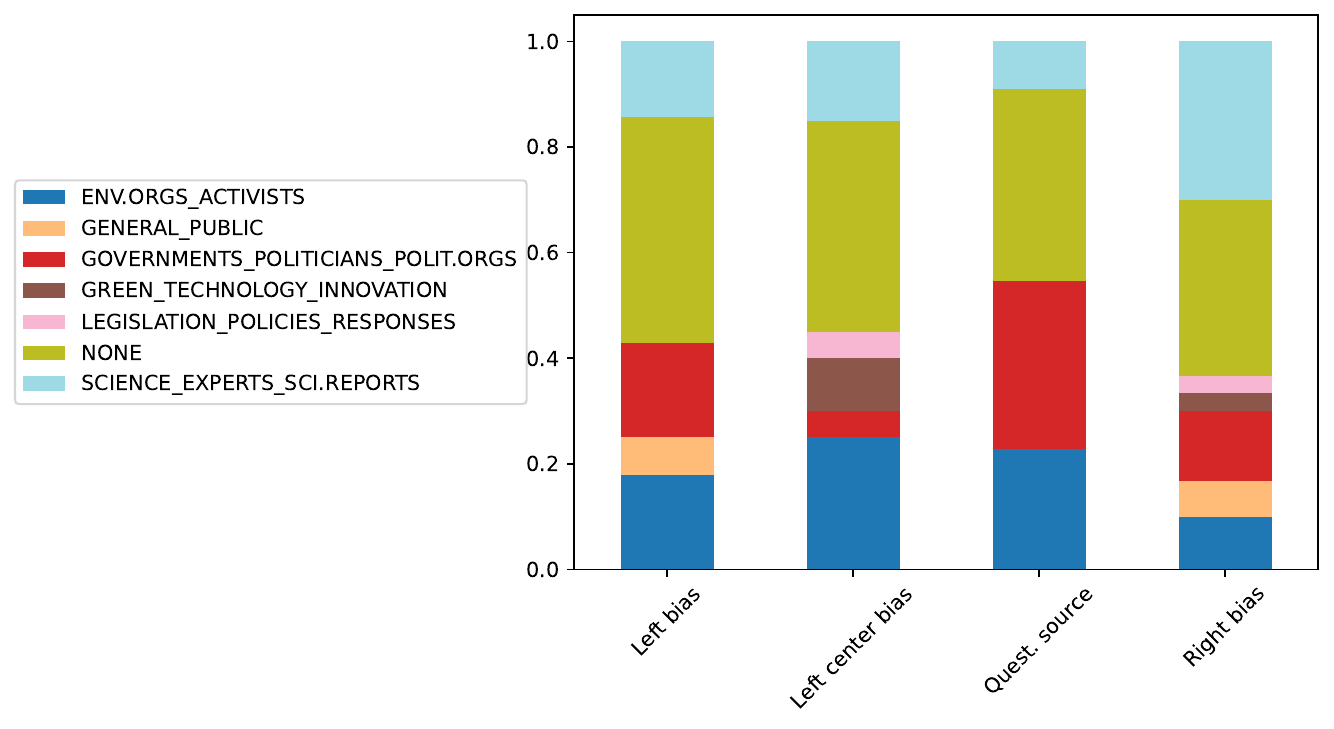}
    \caption{Distribution of entities representing HERO across political leanings}
    \label{fig:hero_leaning}
\end{figure}

\begin{figure}[h!]
    \centering
    \includegraphics[width=1\linewidth]{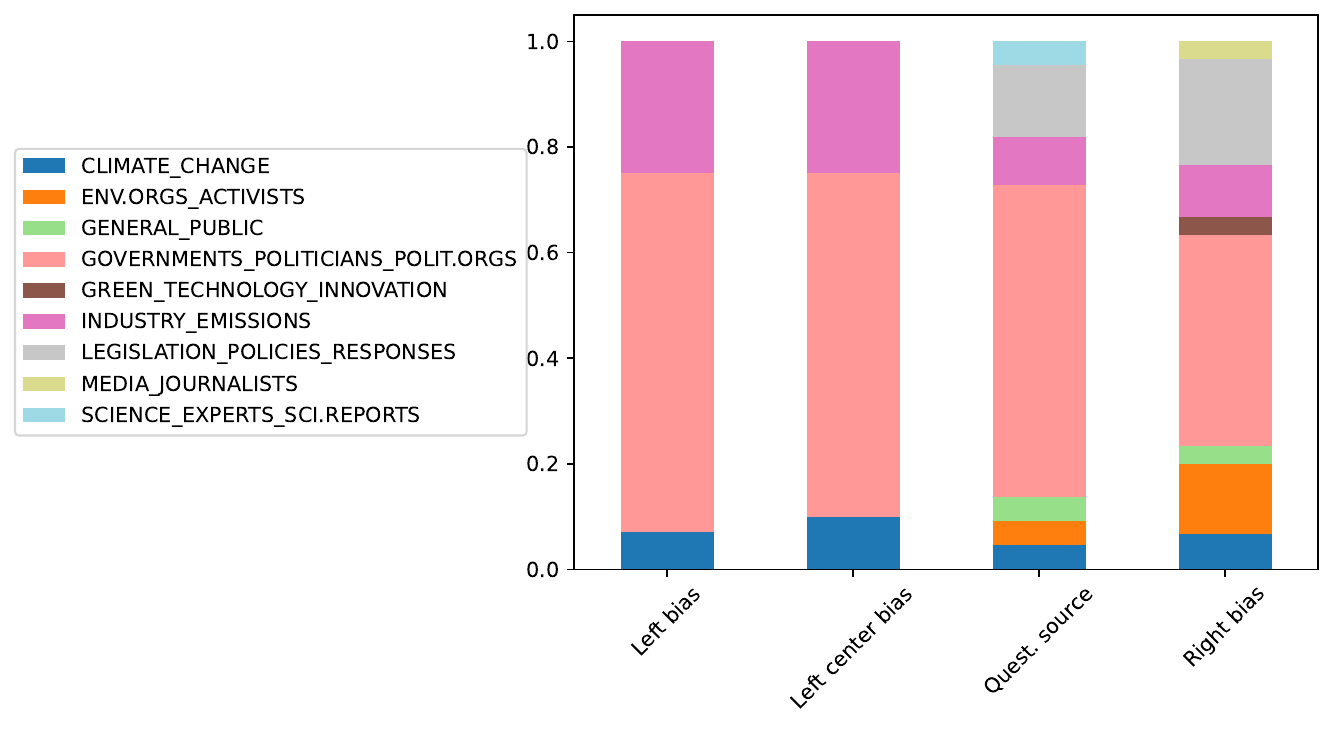}
    \caption{Distribution of entities representing VILLAIN across political leanings}
    \label{fig:villain_leaning}
\end{figure}

\begin{figure}[h!]
    \centering
    \includegraphics[width=1\linewidth]{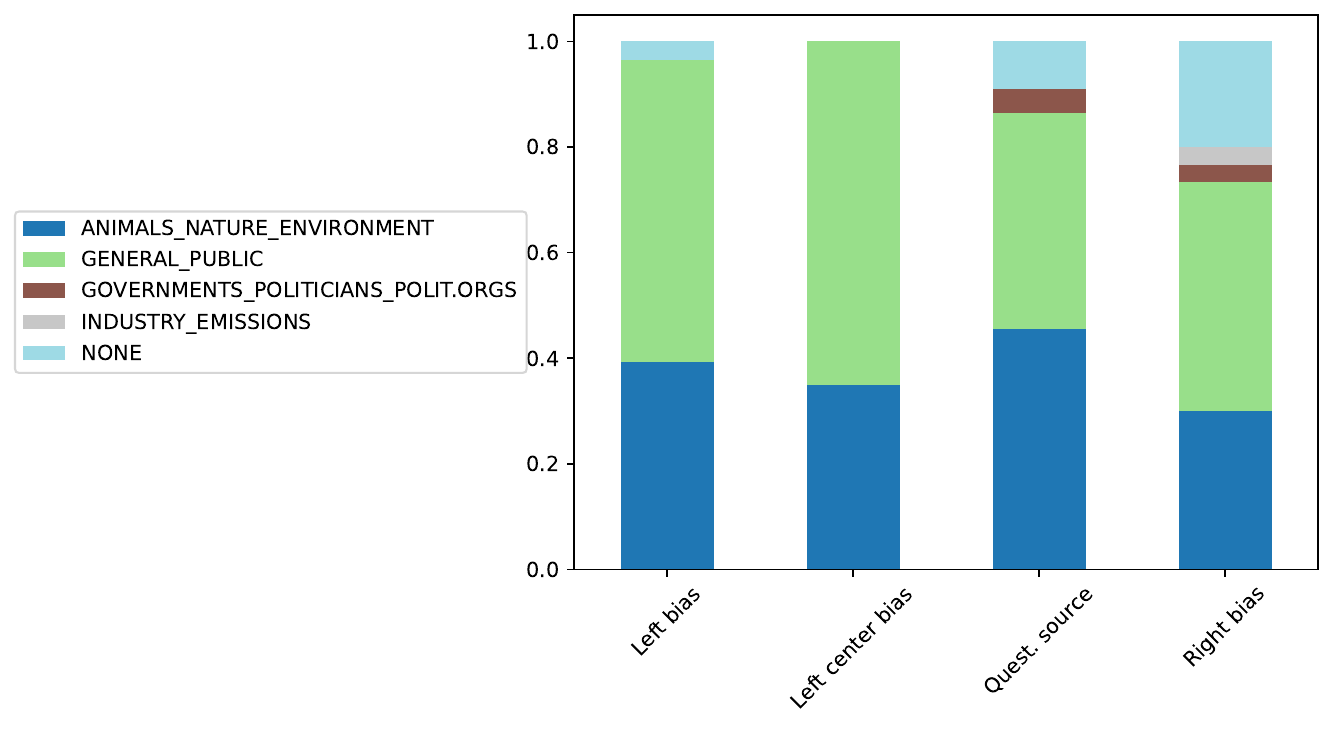}
    \caption{Distribution of entities representing VICTIM across political leanings}
    \label{fig:victim_leaning}
\end{figure}

\begin{figure}[h!]
    \centering
    \includegraphics[width=1\linewidth]{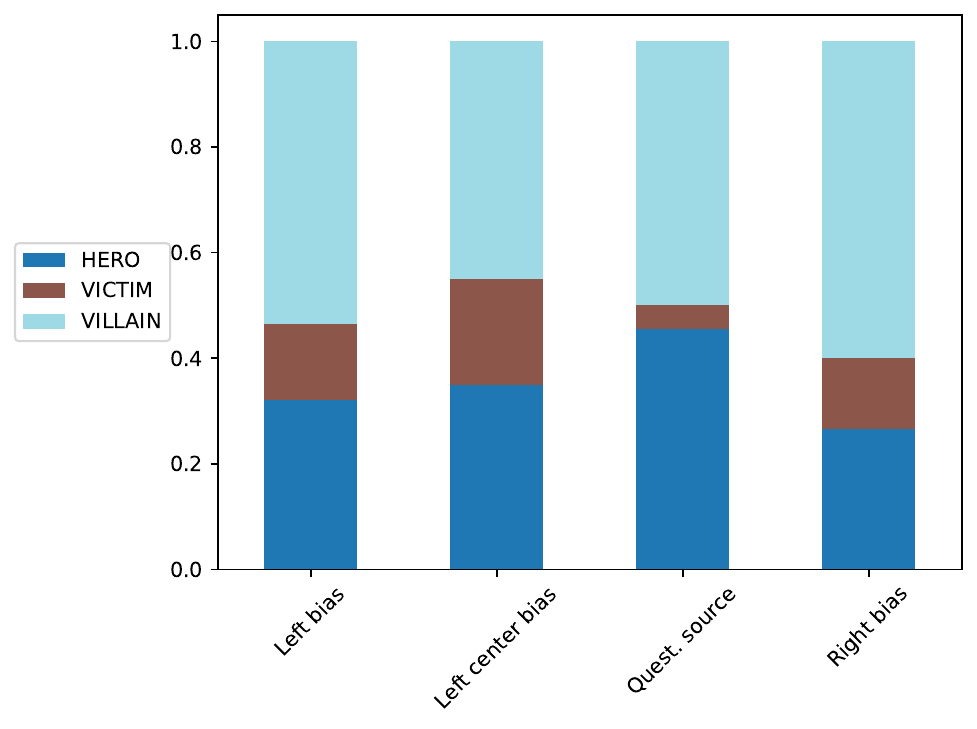}
    \caption{Distribution of FOCUS values across political leanings}
    \label{fig:focus_leaning}
\end{figure}

\begin{figure}[h!]
    \centering
    \includegraphics[width=1\linewidth]{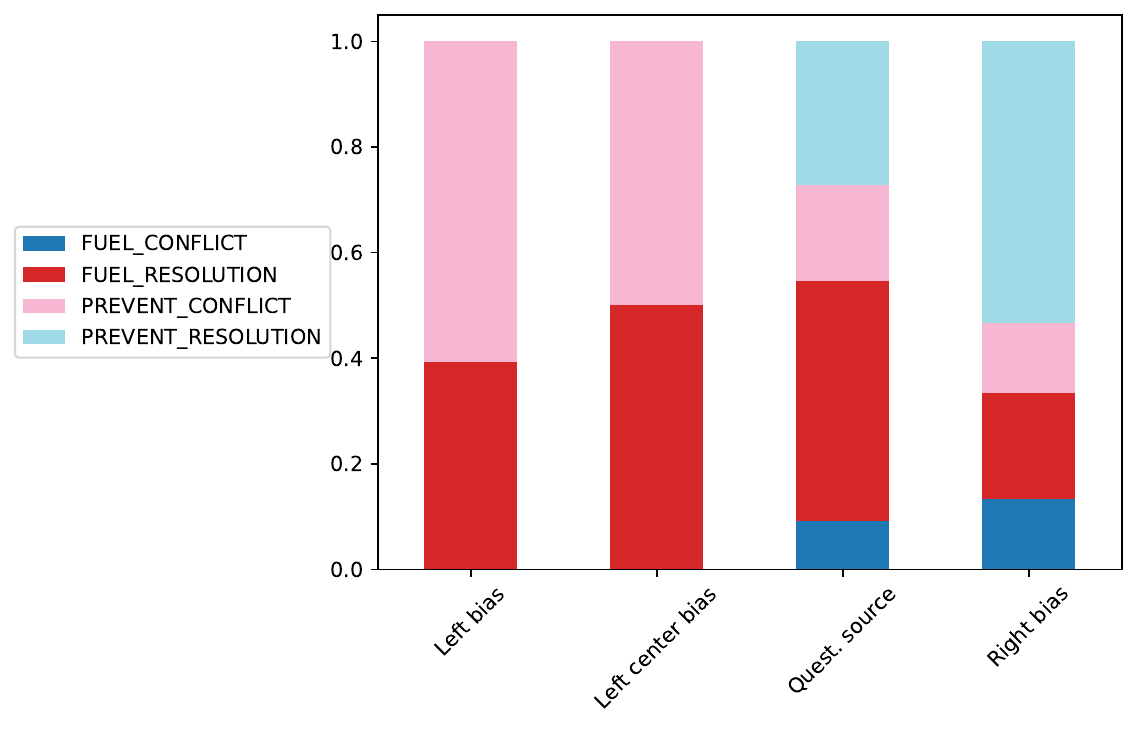}
    \caption{Distribution of CONFLICT values across political leanings}
    \label{fig:conflict_leaning}
\end{figure}

\begin{figure}[h!]
    \centering
    \includegraphics[width=1\linewidth]{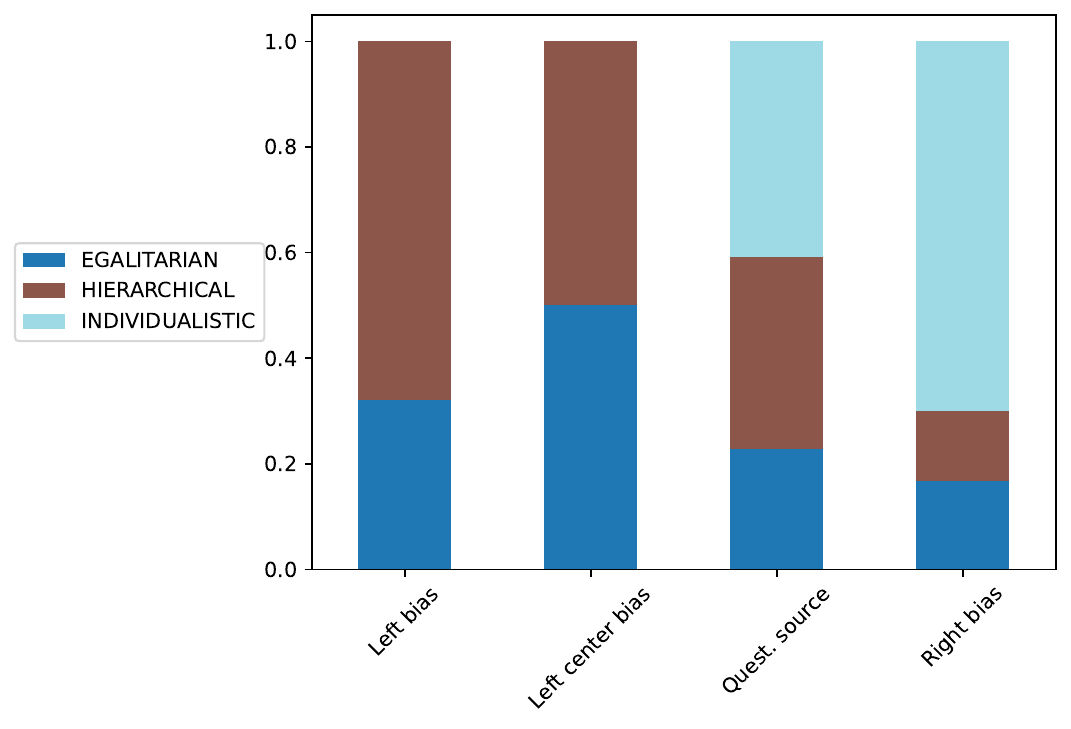}
    \caption{Distribution of CULTURAL STORY values across political leanings\vspace{5ex} }
    \label{fig:story_leaning}
\end{figure}

We explore how different narrative frames and their components are used across political leanings. In particular, we show the distribution of high-level frames (\Cref{fig:narr_leaning}),  narrative frames entities representing \hero (\Cref{fig:hero_leaning}), \villain (\Cref{fig:villain_leaning}), and \victim (\Cref{fig:victim_leaning}); the choice of \focus entity (\Cref{fig:focus_leaning}); the distribution of \conflict values (\Cref{fig:conflict_leaning}) and that of cultural stories 
(\Cref{fig:story_leaning}). Selected analyses are discussed in more detail in the main paper \Cref{ssec:analysis}.

\newpage \pagebreak \cleardoublepage

\begin{table*}[!htbp]
\begin{small}
    \begin{center}
    \begin{tabular}{l|ccccccc}
         & Hero & Villain & Victim & Focus& Conflict & Story & Narrative \\
         \toprule
         GPT4o zero-shot &  0.325 & 	0.454	& 0.266	& 0.656	& 0.332	& 0.574	& 0.258\\
         GPT4o 5-shot & 0.414 &	0.357 &	0.319 &	0.613	 & 0.272	& 0.390 &	0.190 \\
        GPT4o 5-shot with CoT & 0.417	& 0.412	 & 0.330	& 0.627 &	0.332	 &0.430	 & 0.178 \\\bottomrule
    \end{tabular}
    \end{center}
    \end{small}
    \caption{Macro-averaged F1 performance of GPT4o with 5 shot prompting and Dspy optimization for 7 narrative frame understanding tasks}
    \label{tab:gpt_comparison}

\end{table*}

\begin{table*}[!htbp]
\begin{small}
    \begin{center}
    \begin{tabular}{l|ccccccc}
         & Hero & Villain & Victim & Focus& Action & Story & Narrative \\
         \toprule
         Without LoRA &  0.271	& \textbf{0.156}	& \textbf{0.336}	& \textbf{0.568}	& \textbf{0.379}	& \textbf{0.449} & \textbf{0.181}\\
        With LoRA & \textbf{0.338} &	0.118	& 0.221	& 0.351	& 0.231	& 0.393	& 0.077 \\\midrule
         \bottomrule
    \end{tabular}
    \end{center}
    \end{small}
    \caption{Macro-averaged F1 performance of Llama 3.1 with vs without LoRA fine-tuning for 7 narrative understanding tasks}
    \label{tab:lora}

\end{table*}

\section{Additional experiment details}
\label{app:add_exp}

We examine if the performance can be improved by exposing models to annotated examples and optimizing the prompts by adding Chain-of-Thought steps. First, we use 5 randomly selected samples from our dataset for 5-shot learning with GPT4o model. However, except for Hero stakeholder identification, where it leads to some gains, it causes overgeneralization to seen labels and thus drop in performance (see \Cref{tab:gpt_comparison}). We observe similar effects when we perform Low-Rank Adaption (LoRA) fine-tuning \citep{hu2021lora} of Llama.\footnote{We choose Llama as a stronger model among open-source ones, and perform 5-fold fine-tuning with 20\% holdout set, ensuring balanced class representation (hyperparameters and details in \Cref{app:size}): despite improved classification of Hero, the overall performance drops (\Cref{tab:lora}).} Similarly, we notice that the fine-tuned model tends to overpredict the most prominent labels, discarding minor classes.

We also use the 5 random samples for a DSPy program \citep{khattab2023dspy} to automatically generate and optimize reasoning steps for Chain-of-Thought (CoT) prompting. The gains (compared to non-optimized 5-shot prompting) are also minimal (see \Cref{tab:gpt_comparison}). In addition, we tried implementing Chain-of-Thought (CoT) manually for HVV identification tasks, where we guide the model through the steps of identifying candidate entities, choosing most prominent among them, and finally classifying their stakeholder type, but this lead to worse performance.

Overall, these additional experiments show that the tasks are difficult to meaningfully learn from examples or even through reasoning steps.

%\begin{figure}
 %   \centering
  %  \includegraphics[width=0.8\linewidth]{5shot.pdf}
   % \caption{Performance of GPT-4 with 0- and 5-shot prompting and 5-shot CoT prompting optimized using DSPy as macro-averaged F1.}
   % \label{fig:five_shot}
%\end{figure}

\section{Narrative frame prediction with and without structure}
\label{app:confusion}
\begin{figure}[h]
    \centering
    \includegraphics[width=1\linewidth]{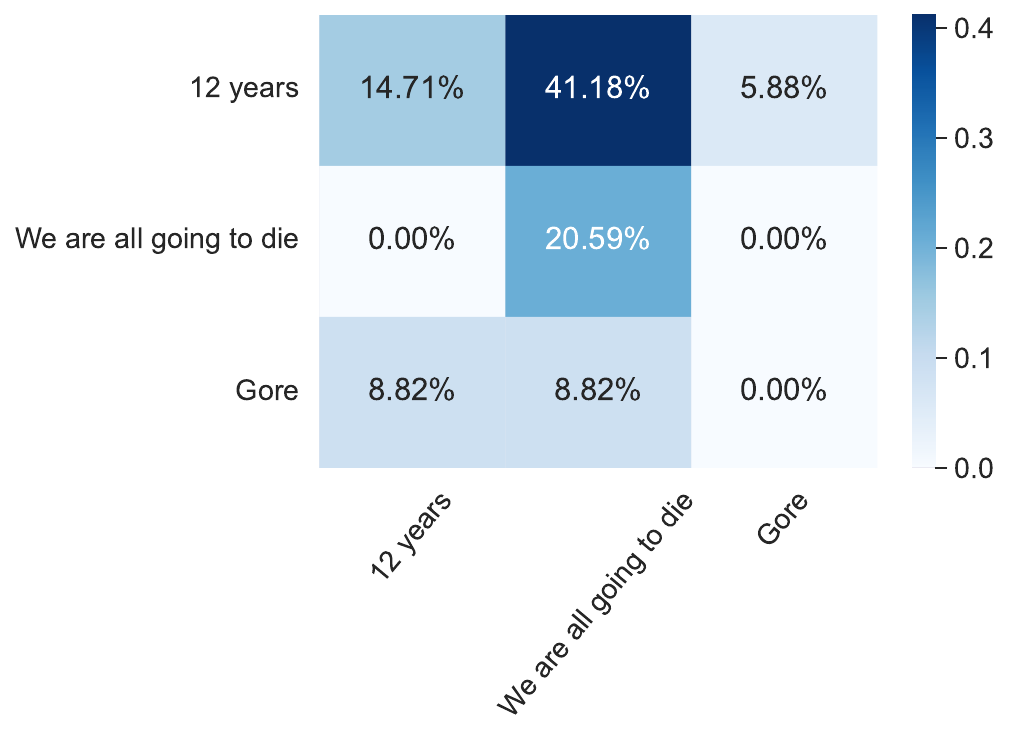}
    \caption{Confusion matrix for zeroshot prediction of only 3 narratives with GPT-4.}
    \label{fig:confusion_matrix}
\end{figure}

\begin{figure*}
    \centering
    \includegraphics[width=1\linewidth]{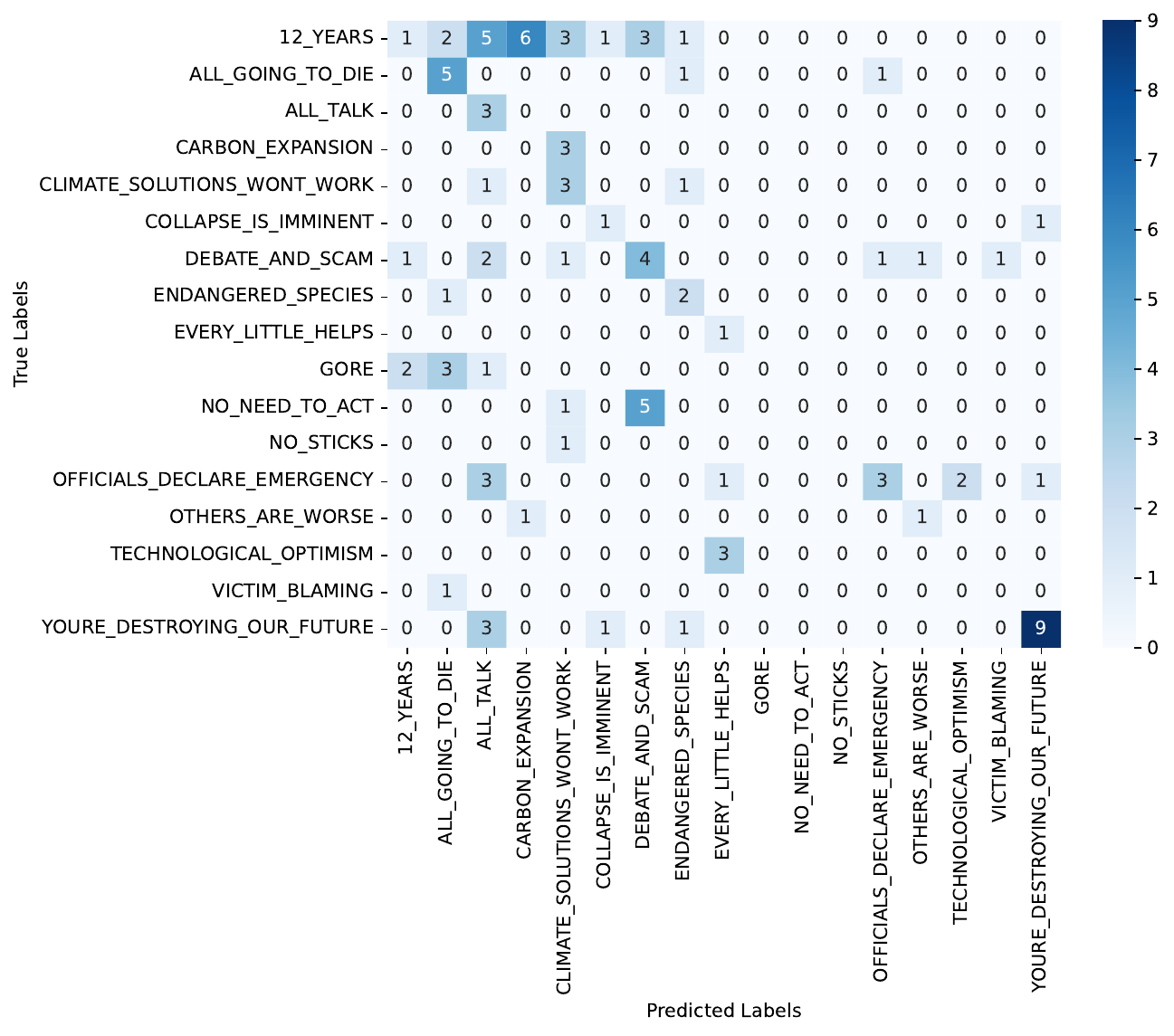}
    \caption{Confusion matrix for Narrative frames prediction using the basic prompt}
    \label{fig:cm_old}
\end{figure*}

\begin{figure*}
    \centering
    \includegraphics[width=1\linewidth]{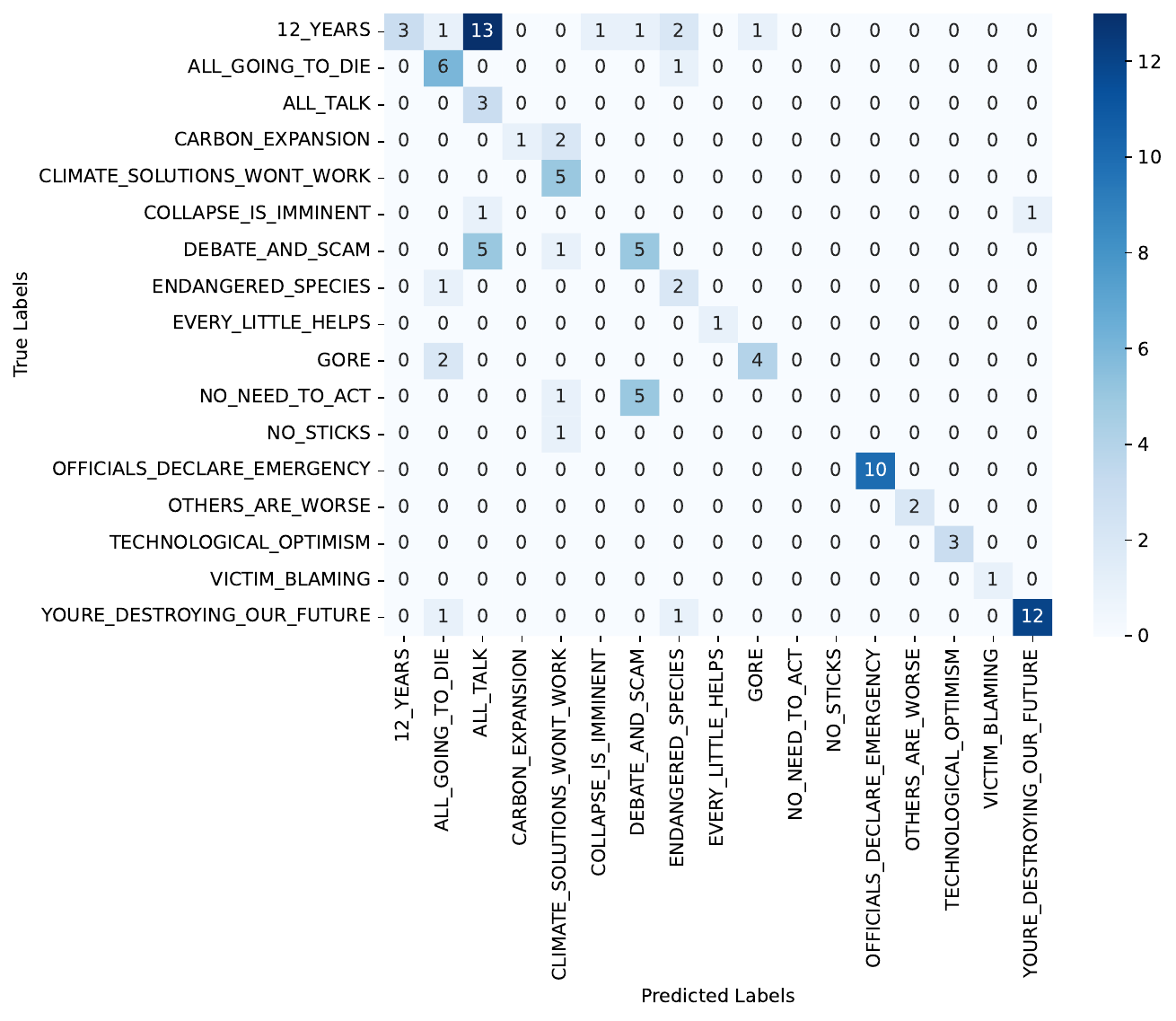}
    \caption{Confusion matrix for Narrative frames prediction using the structured prompt with oracle labels}
    \label{fig:cm_new}
\end{figure*}

Below we show confusion matrices for GPT4o with a basic prompt (\Cref{fig:cm_old}) vs with a structured prompt and oracle (human-annotated) (\Cref{fig:cm_new}) labels.

\section{Prompts}
\subsection{Basic prompts}
\label{app:prompts}

In the tables below we show the basic prompts used for the classification: \Cref{tab:hvv_prompt} for Hero, Villain, Victim and Focus classes, \Cref{tab:conflict_prompt} for Conflict and resolution classification, \Cref{tab:story_prompt} for Story classes, and \Cref{tab:narrative_prompt} for Narrative frame classification.

\begin{table*}[h]

    \begin{tabular}{p{\textwidth}}
         \toprule
You are a social scientist specializing in climate change. You will be given a newspaper article and asked who is framed as a hero, villain or a victim in it.\\

For each of these categories, you will be also asked to specify the corresponding word or phrase, and to classify it into the following classes:\\

GOVERNMENTS\_POLITICIANS: governments and political organizations;\\

INDUSTRY\_EMISSIONS: industries, businesses, and the pollution created by them;\\
LEGISLATION\_POLICIES: policies and legislation responses;\\
GENERAL\_PUBLIC: general public, individuals, and society, including their wellbeing, status quo and economy;\\
ANIMALS\_NATURE\_ENVIRONMENT: nature and environment in general or specific species;\\
ENV.ORGS\_ACTIVISTS: climate activists and organizations\\
SCIENCE\_EXPERTS\_SCI.REPORTS: scientists and scientific reports/research\\
CLIMATE\_CHANGE: climate change as a process or consequence\\
GREEN\_TECHNOLOGY\_INNOVATION: innovative and green technologies\\
MEDIA\_JOURNALISTS: media and journalists\\

Finally, you need to detect which of the characters (hero, villain, or victim) the news story is focusing on.\\

Please return a json object which consists of the following fields:\\

hero\_class: a label for the hero from the list above, or 'None' if the hero cannot be identified.\\
villain\_class: a label for the villain from the list above, or 'None' if the villain cannot be identified.\\
victim\_class: a label for the victim from the list above, or 'None' if the victim cannot be identified.\\
focus: one of the following - HERO, VILLAIN, VICTIM\\

         \bottomrule

    \end{tabular}
    \caption{Basic prompt for Hero, Villain, Victim, and Focus classification}
    \label{tab:hvv_prompt}
\end{table*}

\begin{table*}[t]

    \begin{tabular}{p{\textwidth}}
         \toprule
You are a social scientist specializing in climate change.\\
You will be given a newspaper article and asked to identify how it relates to climate crisis.\\
Assign one of the following classes:\\

FUEL\_RESOLUTION: the article proposes or describes specific measures, policies, or events that would contribute to the resolution of the climate crisis.\\
FUEL\_CONFLICT: the article proposes or describes specific measures, policies, or events that would exacerbate the climate crisis.\\
PREVENT\_RESOLUTION: the article criticises measures, policies, or events that contribute to the resolution of the climate crisis; or it denies the climate crisis.\\
PREVENT\_CONFLICT: the article criticises measures, policies, or events that exacerbate the climate crisis; or it provides the evidence for the climate crisis.\\

Please return a json object which consists of the following field:\\

conflict: one of the following labels: FUEL\_RESOLUTION, FUEL\_CONFLICT, \\PREVENT\_RESOLUTION, PREVENT\_CONFLICT.\\
\bottomrule
\end{tabular}
    \caption{Basic prompt for Conflict classification}
    \label{tab:conflict_prompt}
\end{table*}

\begin{table*}[ht]

    \begin{tabular}{p{\textwidth}}
         \toprule
You are a social scientist specializing in climate change.\\
You will be given a newspaper article and asked what is the cultural story reflected in it.\\
You should choose one of the following classes:\\

HIERARCHICAL: this story assumes that the situation can be controlled externally, but we need to be bound by tight social prescriptions and group actions.\\
INDIVIDUALISTIC: this story assumes that the situation cannot be controlled externally, and no group actions are necessary.\\
EGALITARIAN: this story assumes that the situation requires combined efforts and group actions of all members of society.\\

Please return a json object which consists of the following field:\\

story: a label from the classes above.\\
\bottomrule
\end{tabular}
    \caption{Basic prompt for Cultural story classification}
    \label{tab:story_prompt}
\end{table*}

\begin{table*}
\begin{tabular}{p{\textwidth}}
         \toprule
You are a social scientist specializing in climate change.\\
You will be given a newspaper article and asked what is the main narrative in it.\\
You should choose one of the following classes:\\

12\_YEARS: 12 Years to save the world - Past and present human action (or inaction) risks a catastrophic future climatic event unless people change their behaviour to mitigate climate change.\\
ALL\_GOING\_TO\_DIE:	We are all going to die - This narrative shows the current or potential catastrophic impact of climate change on people\\
ALL\_TALK:	All talk little action - This narrative emphasises inconcistency between ambitious climate action targets and actual actions.\\
CARBON\_EXPANSION:	Carbon-fuelled expansion - The free market is at the centre of this narrative which presents action on climate change as an obstacle to the freedom and well-being of citizens.\\
CLIMATE\_SOLUTIONS\_WONT\_WORK: Climate solutions won’t work. Climate policies are harmful and a threat to society and the economy. Climate policies are ineffective and too difficult to implement.\\
COLLAPSE\_IS\_IMMINENT: The climate crisis is due to the inaction of the negligent or complacent politicians, and it is incumbent upon individuals to shock society into urgent action\\
DEBATE\_AND\_SCAM: The heroes of this narrative are sceptical individuals who dare to challenge the false consensus on climate change which is propagated by those with vested interests.\\
ENDANGERED\_SPECIES: Endangered species (like polar bears) are the helpless victims of this narrative, who are seeing their habitat destroyed by the actions of villainous humans.\\
EVERY\_LITTLE\_HELPS:	This narrative presents a society which has transitioned to a sustainable ‘green’ way of life. Could be by portraying individuals as the protagonists of stories that propose solutions to climate change.\\
GORE: This is a narrative of scientific discovery which climaxes on the certainty that climate change is unequivocally caused by humans.\\
NO\_STICKS:	No sticks just carrots -  Society will only respond to supportive and voluntary policies, restrictive measures will fail and should be abandoned.\\
OFFICIALS\_DECLARE\_EMERGENCY: Officials declare a climate emergency - The climate crisis is sufficiently severe that it warrants declaring a climate emergency. This should occur at different levels of government as climate requires action at all levels, from the hyper-local to the global.\\
OTHERS\_ARE\_WORSE:	Others are worse than us - Other countries, cities or industries are worse than ourselves. There is no point for us to  implement climate policies, because we only cause a small fraction of the emissions. As long as others emit even more than us, actions won’t be effective.\\
TECHNOLOGICAL\_OPTIMISM:	We should focus our efforts on current and future technologies, which will unlock great possibilities for addressing climate change.\\
VICTIM\_BLAMING:	Individuals and consumers are ultimately responsible for taking actions to address climate change.\\
YOURE\_DESTROYING\_OUR\_FUTURE: The political stasis around climate change means that we cannot rely on politicians to create the change necessary. With collective action, even the politically weak can make a difference and secure a future for generations to come.\\
Please return a json object which consists of the following field:\\
narrative: a label from the classes above.\\

\bottomrule

    \end{tabular}
    \caption{Basic prompt for Narrative classification}
    \label{tab:narrative_prompt}
\end{table*}

\subsection{Modified prompts with structure descriptions}
\label{app:mod_prompts}

In \Cref{tab:narrative_str_prompt} below we show the modified prompts used for Narrative prediction.

\begin{table*}[ht]
\begin{tabular}{p{\textwidth}}
         \toprule
You are a social scientist specializing in climate change.\\
You will be given a newspaper article and asked what is the main narrative in it.
You should choose one of the following classes:\\
\\
12\_YEARS: 12 Years to save the world - Past and present human action (or inaction) risks a catastrophic future climatic event unless people change their behaviour to mitigate climate change. The villain here is government or industry pollution, and the victim is environment, people, or climate change. The narratives focuses on villain and shows how they deny climate change or abandon climate policies.\\
ALL\_GOING\_TO\_DIE:	We are all going to die - This narrative shows the current or potential catastrophic impact of climate change on people. The villain here is climate change or industry emissions, and the victim is general public. The narrative focuses on victim and raises the alarm.\\
ALL\_TALK:	All talk little action - This narrative emphasises inconcistency between ambitious climate action targets and actual actions. The villain here is government and politicians, and the victim is often climate change. The narrative focuses on villain who reneged on their promise to support climate policies.\\
CARBON\_EXPANSION:	Carbon-fuelled expansion - The free market is at the centre of this narrative which presents action on climate change as an obstacle to the freedom and well-being of citizens. The villain here is climate policies or green technologies, and the victim is general public or old industries. The narrative focuses on victim and advocates for abandoning climate policies.\\
CLIMATE\_SOLUTIONS\_WONT\_WORK: Climate solutions won't work. Climate policies are harmful and a threat to society and the economy. Climate policies are ineffective and too difficult to implement. The villain is here climate policies or green technologies, and the victim is usually general public. The narrative focuses on villain and criticizes them.\\
COLLAPSE\_IS\_IMMINENT: The climate crisis is due to the inaction of the negligent or complacent politicians, and it is incumbent upon individuals to shock society into urgent action. The heroes here are environmental activists, and the villain is government. The narrative focuses on villain and advocated for taking action such as protests or disobedience.\\
DEBATE\_AND\_SCAM: The heroes of this narrative are sceptical individuals who dare to challenge the false consensus on climate change which is propagated by those with vested interests. The villains are  governments, activists, journalist and policies that support climate measures. The narrative focuses on villains and exposes them.\\
ENDANGERED\_SPECIES: Endangered species (like polar bears) are the helpless victims of this narrative, who are seeing their habitat destroyed by the actions of villainous humans. The villain here can be government, legislation, industry, and the victim is environment and nature. The narrative focuses on victims and shows how they are endangered.\\
EVERY\_LITTLE\_HELPS:	This narrative presents a society which has transitioned to a sustainable ‘green’ way of life. Could be by portraying individuals as the protagonists of stories that propose solutions to climate change. The heroes here are individuals and common people, and it is implied that they are also a villain. The narrative focuses on hero and shows how they change their consumption.\\
GORE: This is a narrative of scientific discovery which climaxes on the certainty that climate change is unequivocally caused by humans. The heroes here are scientists, the villain is government, general public, or industry pollution, and the victim is environment or climate change. The narrative focuses on villain and raises alarm.\\
...
\\
Please return a json object which consists of the following field:\\

narrative: a label from the classes above.\\
\bottomrule

    \end{tabular}
    \caption{Prompt for Narrative classification with Hero, Villain, Victim, and Focus specified (abbreviated)}
    \label{tab:narrative_str_prompt}
\end{table*}

\newpage

\section{COVID-19: HVV stakeholder extraction}
\label{app:stakeholder_generation}

In this sections we provide prompts we used for multi-step clustering and extraction of stakeholder classes, and well as the list of the resulting classes to be used in HVV classification prompts.

\subsection{Prompts}

We provide prompts for identifying candidate entities in each speech \Cref{tab:hvv_1}, and then clustering them into stakeholder types \Cref{tab:hvv_2}.

\begin{table*}[h]
\begin{tabular}{p{\textwidth}}
         \toprule
You are a social scientist specializing in media analysis. You will be given a politician's address and asked asked who or what is framed as a hero, villain or a victim in it.\\
    List the entities corresponding to these character roles, and cluster them according to their type (i.e.~what kind of entity they represent).\\
    Please return a json object which consists of the following fields:\\
    heroes: a list of entity types that you identified as heroes,\\
    villains: a list of entity types that you identified as villains,\\
    victims: a list of entity types that you identified as victims.\\

    Do not include anything apart from these fields.\\

\bottomrule

    \end{tabular}
    \caption{Basic prompt for candidate characters extraction}
    \label{tab:hvv_1}
\end{table*}

\begin{table*}[h]
\begin{tabular}{p{\textwidth}}
         \toprule
You are a social scientist specializing in media analysis. You will be given a list of entities that appear in politicians speeches regarding Covid 19.\\
Many of these entities are similar or overlapping. Cluster them to derive the main actors or stakeholders groups. \\
\bottomrule

    \end{tabular}
    \caption{Basic prompt for grouping entities into stakeholder types}
    \label{tab:hvv_2}
\end{table*}

\subsection{Resulting classes}

\begin{itemize}
    \item HEALTHCARE: frontline workers, medical professionals, and institutions directly involved in providing care and combatting the pandemic;
    \item VULNERABLE\_POPULATION: individuals at higher risk of severe illness or death from COVID-19;
    \item GENERAL\_PUBLIC: general public, individuals, communities, and society;
    \item GOVERNMENT\_POLITICIANS: national and regional governments and policymakers;
    \item BUSINESS\_ECONOMY: businesses, workers, and the broader economy;
    \item SCIENCE\_EXPERTS: scientists, researchers, and research institutions;
    \item FAITH\_GROUPS: faith-based organizations;
    \item PANDEMIC: the virus itself and the pandemic;
    \item GLOBAL\_EFFORTS: international organizations, global collaborations, and efforts to address the pandemic on a worldwide scale.
 
\end{itemize}

\end{document}